\documentclass[11pt]{article}

\usepackage[final]{acl}

\usepackage{times}
\usepackage{latexsym}

\usepackage[T1]{fontenc}

\usepackage[utf8]{inputenc}

\usepackage{microtype}

\usepackage{inconsolata}

\usepackage{graphicx}
\usepackage{booktabs}
\usepackage{subcaption}
\usepackage{wrapfig}
 \usepackage{multirow} 
\usepackage{listings}
\usepackage{xcolor}
\usepackage{inconsolata}
\usepackage{makecell}

\definecolor{codegreen}{rgb}{0,0.6,0}
\definecolor{codegray}{rgb}{0.5,0.5,0.5}
\definecolor{codepurple}{rgb}{24, 24, 245}
\definecolor{backcolour}{rgb}{255, 255, 255}

\definecolor{taskblue}{RGB}{0,99,177}
\definecolor{refgreen}{RGB}{0,150,85}
\definecolor{reforange}{RGB}{241, 157, 56}
\definecolor{refpurple}{RGB}{159, 77, 182}
\definecolor{refblue}{RGB}{77, 161, 255}

\definecolor{subviolet}{RGB}{131,76,190}
\definecolor{templateblue}{RGB}{1, 128, 134}
\definecolor{refgreenDark}{RGB}{0, 90, 45}
\definecolor{analysisblue}{HTML}{1E90FF} 
\definecolor{algoyellow}{HTML}{A0522D}   

\lstdefinestyle{mystyle}{
    backgroundcolor=\color{backcolour},   
    commentstyle=\color{codegreen},
    keywordstyle=\color{magenta},
    numberstyle=\scriptsize\color{codegray},
    stringstyle=\color{codepurple},
    basicstyle=\ttfamily\scriptsize,
    breakatwhitespace=false,         
    breaklines=true,                 
    captionpos=b,                    
    keepspaces=true,                 
    numbers=left,                    
    numbersep=5pt,                  
    showspaces=false,                
    showstringspaces=false,
    showtabs=false,                  
    tabsize=2,
    frame=single,
    rulecolor=\color{black},
    numbers=none  
}

\lstdefinestyle{codestyle}{
  language=Python,
  basicstyle=\footnotesize\ttfamily,
  frame=single,
  numbers=left,
  numberstyle=\scriptsize,
  xleftmargin=1.5em,
  framexleftmargin=1.5em,
  keywordstyle=\color{taskblue},
  commentstyle=\itshape\color{gray},
  stringstyle=\color{orange},
  showstringspaces=false,
  breaklines=true,
  tabsize=2
}

\definecolor{vscode-bg}{RGB}{248,248,248}
\definecolor{vscode-border}{RGB}{220,220,220}
\definecolor{vscode-keyword}{RGB}{0,122,204}
\definecolor{vscode-inline}{RGB}{230,150,50}
\definecolor{vscode-list}{RGB}{150,150,150}
\definecolor{vscode-quote}{RGB}{120,120,120}
\definecolor{vscode-text}{RGB}{40,40,40}

\lstdefinelanguage{Markdown}{
  morecomment=[l][\bfseries\color{vscode-keyword}]{\#},  
}

\lstdefinestyle{mdstyle}{
  language=Markdown,
  basicstyle=\ttfamily\scriptsize\color{vscode-text},
  backgroundcolor=\color{vscode-bg},
  frame=single,
  rulecolor=\color{vscode-border},
  showstringspaces=false,
  breaklines=true,
  columns=fullflexible,
  breakindent=0pt,        
  aboveskip=6pt,
  belowskip=6pt,
  xleftmargin=0pt,
  xrightmargin=0pt,
  framexleftmargin=0pt,
  framexrightmargin=0pt,
  literate=
    {>}{{\textcolor{vscode-quote}{>}}}1
    {-}{{\textcolor{vscode-list}{-}}}1
    {`}{{\textcolor{vscode-inline}{`}}}1
}

\usepackage{array}          
\usepackage{tabularx}       
\usepackage{verbatim}       
\usepackage{fancyvrb}       
\usepackage{listings}       
\usepackage{enumitem}       
\usepackage{amsmath}        
\usepackage{geometry}       

\usepackage{booktabs}       
\usepackage{float}          
\usepackage{caption}        

\usepackage{soul}
\usepackage{xcolor}
\usepackage{colortbl}    

\usepackage[ruled,vlined,linesnumbered]{algorithm2e}
\usepackage[most]{tcolorbox}
\usepackage{comment}
\definecolor{bggreen}{RGB}{237, 237, 237}
\lstdefinelanguage{json}{
    basicstyle=\ttfamily\small,
    numbers=left,
    numberstyle=\tiny\color{gray},
    stepnumber=1,
    numbersep=8pt,
    showstringspaces=false,
    breaklines=true,
    frame=lines,
    backgroundcolor=\color{gray!5},
    stringstyle=\color{red},
    keywordstyle=\color{blue},
    commentstyle=\color{green!50!black}\itshape,
    morekeywords={true, false, null}
}

\title{
\Large\textbf{AscendKernelGen: A Systematic Study of LLM-Based Kernel Generation for Neural Processing Units}\\[0.6ex]
\small\textit{System Design, Data Construction, Fine-Tuning, and Comprehensive Evaluation}
}

\author{
 \textbf{Xinzi Cao\textsuperscript{1,3}\thanks{Co-first author}},
 \textbf{Jianyang Zhai\textsuperscript{1,3}\footnotemark[1]},
 \textbf{Pengfei Li\textsuperscript{2}\footnotemark[1]},
 \textbf{Zhiheng Hu\textsuperscript{2}\footnotemark[1]},
 \textbf{Cen Yan\textsuperscript{2}},
 \\
 \textbf{Bingxu Mu\textsuperscript{2}},
 \textbf{Guanghuan Fang\textsuperscript{2}},
 \textbf{Bin She\textsuperscript{2}},
 \textbf{Jiayu Li\textsuperscript{2}},
 \textbf{Yihan Su\textsuperscript{2}}, 
 \textbf{Dongyang Tao\textsuperscript{2}},
 \\
 \textbf{Xiansong Huang\textsuperscript{1}},
 \textbf{Fan Xu\textsuperscript{1}},
 \textbf{Feidiao Yang\textsuperscript{1}},
 \textbf{Yao Lu\textsuperscript{1}},
 \textbf{Chang-Dong Wang\textsuperscript{3}},
 \textbf{Yutong Lu\textsuperscript{3}},
\\
 \textbf{Weicheng Xue\textsuperscript{1}\thanks{Corresponding author}},
 \textbf{Bin Zhou\textsuperscript{1}\footnotemark[2]},
 \textbf{Yonghong Tian\textsuperscript{1,4}\footnotemark[2]}
\\
\\
 \textsuperscript{1}Pengcheng Laboratory,
 \textsuperscript{2}Huawei,
 \textsuperscript{3}Sun Yat-sen University,
 \textsuperscript{4}Peking University
\\
 \small{
   \textbf{Correspondence:} \href{mailto:xuewch@pcl.ac.cn}{xuewch@pcl.ac.cn}, 
   \href{mailto:senosy@gmail.com}{senosy@gmail.com},
   \href{mailto:yhtian@pku.edu.cn}{yhtian@pku.edu.cn}
 }
}
\begin{document}
\maketitle
\begin{abstract}
To meet the ever-increasing demand for computational efficiency, Neural Processing Units (NPUs) have become critical in modern AI infrastructure. However, unlocking their full potential requires developing high-performance compute kernels using vendor-specific Domain-Specific Languages (DSLs), a task that demands deep hardware expertise and is labor-intensive. While Large Language Models (LLMs) have shown promise in general code generation, they struggle with the strict constraints and scarcity of training data in the NPU domain. Our preliminary study reveals that state-of-the-art general-purpose LLMs fail to generate functional complex kernels for Ascend NPUs, yielding a near-zero success rate. To address these challenges, we propose AscendKernelGen, a generation-evaluation integrated framework for NPU kernel development. We introduce Ascend-CoT, a high-quality dataset incorporating chain-of-thought reasoning derived from real-world kernel implementations, and KernelGen-LM, a domain-adaptive model trained via supervised fine-tuning and reinforcement learning with execution feedback. Furthermore, we design NPUKernelBench, a comprehensive benchmark for assessing compilation, correctness, and performance across varying complexity levels. Experimental results demonstrate that our approach significantly bridges the gap between general LLMs and hardware-specific coding. Specifically, the compilation success rate on complex Level-2 kernels improves from 0\% to 95.5\% (Pass@10), while functional correctness achieves 64.3\% compared to the baseline's complete failure. These results highlight the critical role of domain-specific reasoning and rigorous evaluation in automating accelerator-aware code generation. AscendKernGen is available at \href{https://huggingface.co/AscendKernelGen}{\textcolor{blue}{AscendKernGen}} and \href{https://github.com/weich97/NPUKernelBench}{\textcolor{blue}{NPUKernelBench}}.
\end{abstract}

\section{Introduction}

The rapid advancement of Artificial Intelligence (AI), particularly deep learning, has fundamentally reshaped modern computing architectures. To meet the ever-increasing demand for computational efficiency and throughput, domain-specific accelerators have emerged as a key solution. Among them, Neural Processing Units (NPUs), such as Huawei's Ascend platform~\cite{liao2021ascend, xue2024unlocking, wroblewski2025parallel}, play a critical role in contemporary AI infrastructure by providing high performance for deep learning workloads. However, the practical efficiency of these accelerators is determined not only by hardware capabilities, but also by the quality of the underlying compute kernels. As a result, developing high-performance, hardware-adapted kernels is a prerequisite for fully unlocking the potential of NPU platforms.

Writing such kernels remains extremely challenging. NPU kernel development typically relies on highly specialized, vendor-specific domain-specific languages (DSLs), such as AscendC, which require developers to possess deep expertise in hardware architecture. This includes fine-grained management of hierarchical memory systems (e.g., global versus on-chip memory), carefully designed data tiling strategies, asynchronous pipeline programming for overlapping computation and data movement, and explicit utilization of vector (SIMD) and matrix (Cube) execution units. The resulting learning curve is steep, making manual kernel development time-consuming, costly, and error-prone. Consequently, kernel optimization has become a major bottleneck in the rapid iteration and deployment of AI applications on NPU architectures.

At a higher level, this challenge exposes a fundamental limitation of current code generation approaches: general-purpose programming knowledge is insufficient for generating correct and efficient code in highly specialized, hardware-specific domains. Motivated by the success of Large Language Models (LLMs) in general-purpose code generation, both academia and industry have explored their use for automating kernel development. However, this approach faces a critical obstacle. The knowledge required for hardware-specific DSLs differs fundamentally from that of general-purpose programming languages, encompassing strict API constraints, architecture-dependent semantics, and performance-critical optimization patterns. Moreover, high-quality training data in this domain is extremely scarce, leaving general-purpose LLMs poorly equipped to handle such tasks.

Our preliminary evaluation, summarized in Table~\ref{tab:zero_shot_perf}, confirms this limitation empirically. Even state-of-the-art general-purpose LLMs (e.g., Qwen 3) perform extremely poorly when generating AscendC kernels in a zero-shot setting. These models frequently hallucinate non-existent APIs (e.g., \texttt{AscendC::Softmax}) or severely misuse core hardware interfaces, leading to exceptionally high compilation failure rates. For complex L2/L3 kernels, the execution success rate drops to around 0\%, indicating that, without domain-specific adaptation, general-purpose LLMs are effectively unusable for non-trivial NPU kernel development.

\begin{table}[htbp]
  \centering
  \caption{Zero-shot performance of general-purpose LLMs on the NPUKernelBench benchmark.}
  \label{tab:zero_shot_perf}
  \begin{tabular}{llcc}
    \toprule
    \textbf{Model} & \textbf{Task Type} & \textbf{Compilation} & \textbf{Functional Correctness} \\
     & & \textbf{Success Rate} & \textbf{(Post-Compilation)} \\
    \midrule
    \multirow{2}{*}{Qwen3-8B} & L1 (Simple) & 8.22\% & 1.08\% \\
     & L2/L3 (Complex) & 1.39\% & 0.00\% \\ \hline
     \multirow{2}{*}{Qwen2.5-Coder-7B} & L1 (Simple) & 9.19\% & 0.47\% \\
     & L2/L3 (Complex) & 0.40\% & 0.00\% \\ \hline
     \multirow{2}{*}{Llama3.1-8B} & L1 (Simple) & 23.97\% & 0.69\% \\
     & L2/L3 (Complex) & 19.44\% & 0.00\% \\ \hline
     \multirow{2}{*}{Mistral-7B} & L1 (Simple) & 0.00\% & 0.00\% \\
     & L2/L3 (Complex) & 0.00\% & 0.00\% \\
    \bottomrule
  \end{tabular}
\end{table}

These observations suggest that enabling LLMs to generate high-quality NPU kernels requires two key advancements. First, the model must acquire a deep understanding of the hardware-specific programming paradigm, including API constraints and architectural characteristics. Second, a professional and reliable evaluation framework is necessary to rigorously assess not only compilation success, but also functional correctness and performance, which are all critical for kernel-level code.

To address these challenges, we propose \textbf{AscendKernelGen}, a generation--evaluation integrated framework for NPU kernel development. Our contributions are threefold:

\begin{enumerate}
    \item \textbf{A Domain-Specific Reasoning Dataset for Kernel Generation.}
    We construct and release \textit{Ascend-CoT}, a high-quality dataset designed to capture the structured reasoning processes required for low-level NPU kernel programming. The dataset is curated from real-world kernel implementations and includes detailed annotations that reflect pipeline construction, synchronization logic, and arithmetic reasoning patterns.

    \item \textbf{Domain-Adaptive Post-Training for Low-Level Code Generation.}
    Building on a strong base LLM, we introduce a domain-adaptive post-training strategy that explicitly targets the reasoning challenges of NPU kernel synthesis. The resulting model, \textit{KernelGen-LM}, demonstrates substantially improved compilation success and functional correctness compared to general-purpose code models.

    \item \textbf{A Comprehensive Evaluation Benchmark for NPU Kernels.}
    We design \textit{NPUKernelBench}, an evaluation framework that systematically assesses generated kernels along three dimensions: compilation success, functional correctness, and performance. Its dual-path evaluation design enables rigorous analysis of both static-shape optimization and dynamic-shape robustness, providing a reliable benchmark for future research on accelerator-aware code generation.
\end{enumerate}

Experimental results demonstrate the effectiveness of the proposed approach.
After domain-adaptive post-training on the \texttt{Ascend-CoT} dataset, the model exhibits substantial improvements in kernel generation quality, particularly on complex kernels that are nearly unsolvable in zero-shot settings.
Both compilation success and functional correctness improve consistently across difficulty levels, indicating that domain-specific reasoning supervision is crucial for bridging the gap between general-purpose LLMs and hardware-specific kernel programming.

The remainder of this work is organized as follows.
Section~\ref{sec:relatedworks} reviews prior research on LLM-based code generation, traditional hardware-specific kernel optimization and benchmarks for code generation.
Section~\ref{sec:programming_constraints} introduces the programming abstractions and hardware constraints that characterize NPU kernel development.
Section~\ref{sec:system_overview} presents an overview of the \texttt{AscendKernelGen} framework, covering both the kernel generation pipeline and the hardware-grounded evaluation workflow.
Section~\ref{data_construct} describes the construction of the training data, including documentation-based CoT, code-centric CoT, and general CoT data.
Section~\ref{fine_tune} details the domain-adaptive fine-tuning process for LLM-based NPU kernel generation.
Section~\ref{NPUKernelBench} introduces \texttt{NPUKernelBench}, our evaluation benchmark for assessing compilation, correctness, and performance on real NPU hardware.
Section~\ref{experiments} reports the experimental setup and provides an in-depth analysis of the empirical results.
Finally, Section~\ref{conclusions} concludes the report and discusses directions for future work.
\section{Related Works}
\label{sec:relatedworks}

Our work lies at the intersection of large-scale model–based code generation and hardware-specific kernel optimization. In this section, we review prior work from four perspectives: general-purpose code generation with LLMs, traditional hardware kernel optimization, LLM-based kernel generation, and benchmarking for code generation.

\subsection{Large-Scale Models for General-Purpose Code Generation}

Large Language Models (LLMs) have demonstrated remarkable progress in general-purpose code generation across mainstream programming languages such as Python, C++, and Java. Representative systems include OpenAI's Codex~\cite{chen2021evaluating}, AlphaCode~\cite{novikov2025alphaevolve}, Code Llama~\cite{roziere2023code}, Qwen2.5/3-Coder~\cite{hui2024qwen2}, and CodeRL~\cite{le2022coderl}, enabling applications such as code completion, refactoring, and competitive programming.

Despite these advances, general-purpose LLMs remain limited when applied to domain-specific code generation tasks that involve specialized APIs and strict semantic constraints~\cite{gu2025effectiveness}. In particular, hardware-specific programming languages impose requirements that differ fundamentally from general-purpose software development, including explicit memory hierarchy management, architecture-dependent execution semantics, and performance-critical correctness constraints. When faced with such settings, LLMs often fail to generate compilable or semantically valid code, motivating the need for domain-adaptive data and training strategies.

\subsection{Traditional Hardware-Specific Kernel Optimization}

Prior to the adoption of LLMs, kernel optimization for specialized hardware primarily relied on compiler-based and auto-tuning approaches. Auto-tuning frameworks such as TVM~\cite{chen2018tvm} and Ansor~\cite{zheng2020ansor} search over scheduling and optimization spaces to improve performance across diverse hardware backends. High-level abstractions and domain-specific languages, including Halide~\cite{ragan2013halide}, Triton~\cite{tillet2019triton}, and MLIR~\cite{lattner2020mlir}, decouple algorithm specification from execution scheduling, substantially lowering the barrier to kernel optimization on GPUs and other accelerators. Similarly, TensorFlow XLA~\cite{snider2023operator} and PyTorch 2.0~\cite{ansel2024pytorch} adopt compiler-based pipelines to transform high-level computation graphs into optimized backend-specific kernels.

More specialized approaches leverage polyhedral compilation techniques to model and optimize loop transformations systematically. Representative systems include Tensor Comprehensions~\cite{vasilache2018tensor}, Tiramisu~\cite{baghdadi2019tiramisu}, and AKG~\cite{zhao2021akg}, the latter of which targets NPUs by modeling heterogeneous compute units and complex memory hierarchies. Other efforts explore operator-level search and algebraic transformation, such as TASO~\cite{jia2019taso}, PET~\cite{wang2021pet}, and EINNET~\cite{zheng2023einnet}. In addition, platform-specific libraries like CUTLASS~\cite{markidis2018nvidia} and Liger-Kernel~\cite{hsu2024liger} provide hand-optimized kernels for particular hardware–workload combinations.

While these methods can achieve high performance, they typically require substantial expert knowledge, incur high search or development costs, and offer limited generalization across architectures. General-purpose compilers often lack the flexibility to adapt to closed or interface-constrained platforms, whereas platform-specific libraries sacrifice portability. These limitations motivate exploration of more flexible, learning-based kernel generation approaches.

\subsection{LLM-based Generation for Hardware Kernels}

Inspired by the success of LLMs in general-purpose programming, recent work has begun to explore their application to hardware kernel generation, primarily within mature ecosystems such as CUDA GPUs and TPUs. Early efforts focus on direct kernel generation via prompting and in-context learning, as exemplified by CUDA-LLM~\cite{chen2025cuda} and AI CUDA Engineer~\cite{lange2025ai}. Subsequent work introduces domain adaptation through supervised fine-tuning, such as KernelLLM~\cite{kernelllm2025}, which demonstrates that compact, domain-specific models can generate competitive GPU kernels.

More recent studies incorporate reinforcement learning and execution feedback to iteratively refine generated kernels. Systems such as Kevin~\cite{baronio2025kevin}, CUDA-L1~\cite{li2025cuda}, AutoTriton~\cite{li2025autotriton}, and TritonRL~\cite{woo2025tritonrl} show that combining supervised initialization with execution-guided optimization can significantly improve correctness and performance. In parallel, agentic and multi-agent frameworks have been proposed to automate kernel optimization through planning, debugging, and search, including EvoEngineer~\cite{guo2025evoengineer}, STARK~\cite{dong2025stark}, and CudaForge~\cite{zhang2025cudaforge}. Related efforts also explore retrieval-augmented and graph-based reasoning for kernel optimization~\cite{gong2025large}.

Despite these advances, existing LLM-based kernel generation methods largely focus on open and well-documented platforms such as CUDA and TPU, benefiting from mature toolchains and abundant training data. In contrast, LLM-driven kernel generation for emerging NPU platforms with proprietary programming models remains underexplored.

\subsection{Benchmarking for Code Generation}

Evaluation of LLM-generated code in general-purpose domains has been standardized by benchmarks such as HumanEval~\cite{chen2021evaluating} and MBPP~\cite{austin2021program}. However, these benchmarks focus primarily on functional correctness and are ill-suited for hardware kernel generation, where compilation success, hardware-constrained correctness, and performance metrics are equally critical.

To address this gap, hardware-oriented benchmarks such as KernelBench~\cite{ouyang2025kernelbench} and MultiKernelBench~\cite{wen2025multikernelbench} extend evaluation to include compilation and performance. While these benchmarks represent an important step forward, they typically adopt a single-path evaluation paradigm that requires generation of complete host–device code for all tasks. This design introduces unnecessary complexity for simple kernels and limits flexibility in assessing different levels of kernel abstraction. More recent systems, such as TritonGym~\cite{anonymous2025tritongym}, provide interactive and tool-centric evaluation environments for GPU kernels.

Overall, there remains a lack of a flexible, comprehensive evaluation framework tailored to NPU kernel generation that can jointly assess kernel-only optimization and full host–-device integration. Addressing this gap is essential for reliably measuring progress in LLM-based kernel generation for specialized hardware platforms.
\section{Programming Abstractions and Constraints for NPU Kernel Generation}
\label{sec:programming_constraints}

This section characterizes the programming abstraction underlying low-level NPU kernel development and explains why this abstraction poses fundamental challenges for LLM-based kernel generation. Rather than enumerating hardware-specific details, we focus on the structural properties of kernel programs that jointly determine correctness and performance.

\subsection{A Structured Kernel Programming Abstraction}
\label{sec:kernel_abstraction}

We view a low-level NPU kernel as a statically structured program that jointly specifies global data partitioning, asynchronous pipeline stages, and explicit synchronization semantics under a single replicated execution template.

Under this abstraction, a single kernel program is replicated across multiple processing units following a data-parallel execution model. Each kernel instance operates on a distinct slice of global data determined by a logical block index. As a result, kernel code must explicitly compute global memory offsets, valid data ranges, and boundary conditions, instead of relying on implicit indexing mechanisms.

Computation is further organized as an explicit asynchronous pipeline. Data loading, computation, and write-back are mapped to independent hardware units and are intended to overlap in time. This overlap is not managed by the runtime. Instead, the kernel program must explicitly encode producer--consumer relationships between pipeline stages through synchronization primitives, forming a statically defined execution schedule.

This structure is exposed through low-level programming interfaces that require explicit specification of memory access patterns, strides, masks, and synchronization events. Consequently, arithmetic computation, data movement, and control flow are tightly coupled within the kernel code, rather than being inferred or optimized automatically.

\subsection{Implications for LLM-Based Kernel Generation}
\label{sec:llm_challenges}

The structured nature of this kernel programming abstraction imposes reasoning requirements that go beyond local code synthesis, making low-level NPU kernel generation particularly challenging for Large Language Models.

\paragraph{Long-range semantic dependencies.}
Kernel correctness frequently depends on auxiliary parameters, such as block indices, tiling factors, and boundary sizes, which are computed outside the kernel and passed in at runtime. These parameters are referenced across multiple pipeline stages and simultaneously control memory access and synchronization behavior. Generating correct kernel code therefore requires maintaining semantic consistency across separated code regions, demanding long-range dependency reasoning rather than local pattern completion.

\paragraph{Explicit synchronization reasoning.}
Asynchronous pipeline stages communicate exclusively through manually inserted synchronization primitives. Correct execution relies on precise pairing and ordering of these operations to enforce producer--consumer relationships. Missing, redundant, or misordered synchronization can lead to deadlocks or silent data hazards, making correctness contingent on reasoning about dynamic execution ordering rather than linear instruction sequences.

\paragraph{Boundary-sensitive arithmetic reasoning.}
Because data dimensions are often not aligned with hardware-preferred block sizes, kernel code must explicitly handle boundary cases through offset computation and masking logic. This requires precise arithmetic reasoning over loop bounds, indices, and validity conditions. Small errors in these calculations frequently result in incorrect outputs or invalid memory accesses.

\paragraph{Layout-aware representation reasoning.}
Different pipeline stages may operate on distinct physical data layouts optimized for specific execution units. Kernel code must correctly manage layout transitions while preserving logical tensor semantics. This separation between physical representation and abstract tensor meaning requires reasoning beyond shape-level manipulation and is difficult to infer implicitly.

\section{System Overview} \label{sec:system_overview}

We present \textbf{AscendKernelGen (AKGen)}, a unified research framework for studying Large Language Model (LLM)-based generation of low-level NPU kernels. Rather than serving as a standalone engineering system, AKGen is designed to support systematic investigation of kernel generation, verification, and evaluation under realistic hardware programming constraints.

\begin{figure*}[t]
    \centering
    \includegraphics[width=1\linewidth]{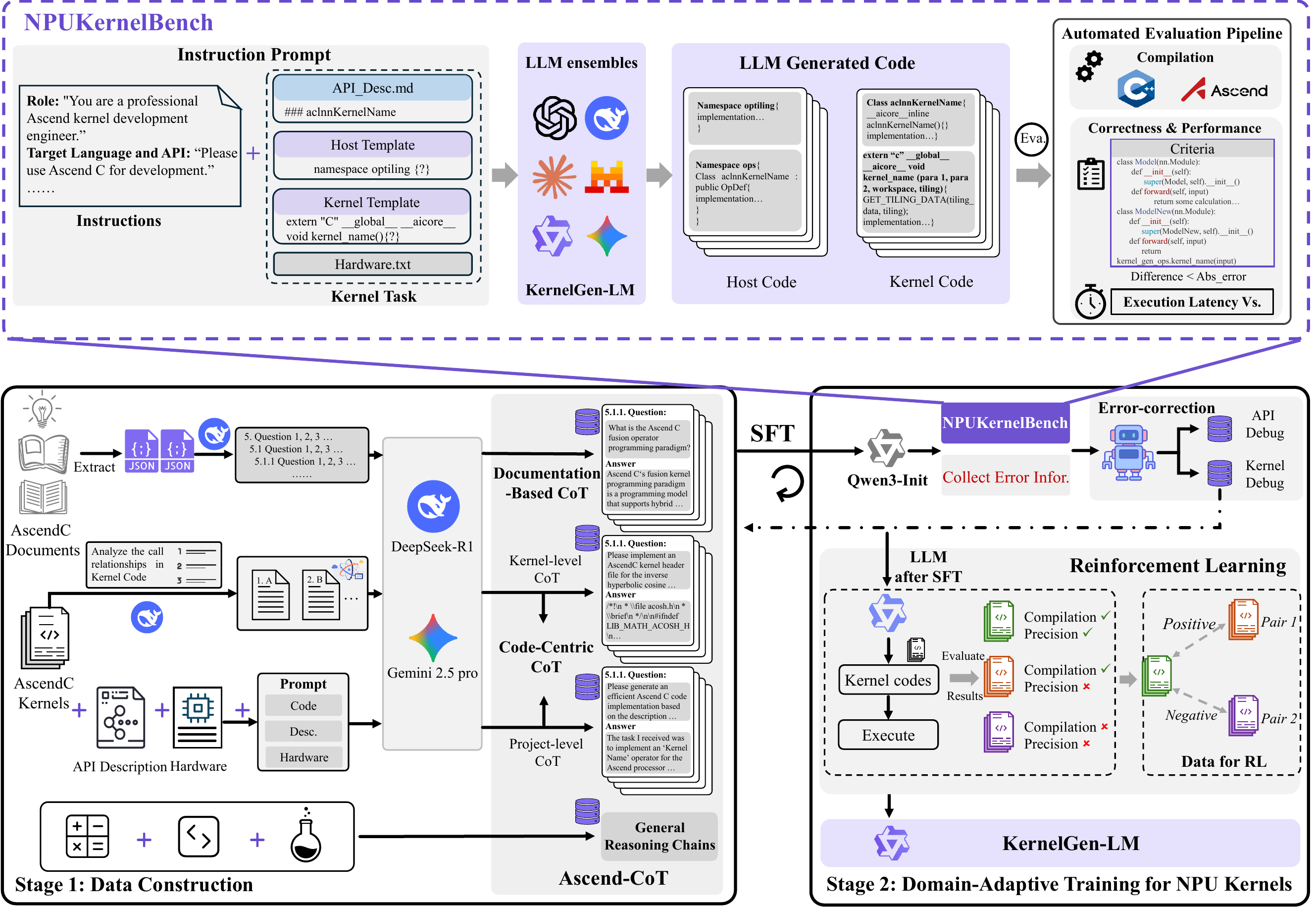}
    \caption{System overview of AscendKernelGen, depicting the data construction, LLM training, and hardware-grounded evaluation pipeline for NPU kernel generation.}
    \label{fig:framework}
\end{figure*}

As illustrated in Fig.~\ref{fig:framework}, AKGen integrates three tightly coupled components: a domain-specific reasoning dataset, a kernel generation model, and a structured evaluation benchmark. Together, they form a closed generation--evaluation loop that enables controlled analysis of model capabilities and limitations.

\paragraph{Generation Component.}
The generation side of AKGen is built upon two elements.

\begin{itemize}
    \item \textbf{Ascend Chain-of-Thought Dataset (Ascend-CoT).}
    To address the scarcity of low-level NPU programming data, we construct Ascend-CoT, a curated dataset that captures the structured reasoning processes required for AscendC kernel development. The dataset is collected from real-world operator implementations and documentation, and emphasizes explicit pipeline construction, synchronization logic, and arithmetic reasoning patterns essential to kernel correctness.

    \item \textbf{KernelGen-LM.}
    Using Ascend-CoT, we perform domain-adaptive post-training on a strong base LLM to obtain \textit{KernelGen-LM}, a model specialized for low-level NPU kernel generation. The training objective explicitly targets the reasoning challenges of hardware-aware code generation, including data tiling, asynchronous pipeline orchestration, and correct use of low-level programming interfaces.
\end{itemize}

\paragraph{Evaluation Component.}
To systematically assess generated kernels, AKGen incorporates a dedicated evaluation framework, \textit{NPUKernelBench}, which bridges the gap between probabilistic model outputs and deterministic hardware execution. NPUKernelBench organizes evaluation tasks into multiple difficulty levels, ranging from simple element-wise operators to complex fused kernels, enabling fine-grained analysis of model behavior across diverse computational patterns and levels of programming complexity.

Each generated kernel is evaluated through a staged validation pipeline that checks compilation success, verifies functional correctness against reference implementations, and measures execution performance in terms of runtime. This multi-stage design isolates distinct failure modes and allows us to quantify model behavior beyond surface-level code validity. While the current study emphasizes correctness and performance metrics, the framework is readily extensible to additional analyses, such as static code quality assessment and deeper performance characterization.

Overall, AKGen provides a controlled framework for analyzing how LLMs acquire, apply, and generalize low-level hardware programming knowledge, with Ascend NPU kernels serving as a representative and challenging testbed.

\section{Data Construction}\label{data_construct}

To equip large language models (LLMs) with the reasoning capabilities required for low-level NPU kernel generation, we construct a multi-source dataset that integrates domain-specific kernel knowledge with general reasoning supervision. The design goal is not only to expose the model to the surface form of AscendC kernels, but also to teach how expert developers reason about tiling strategies, memory movement, API constraints, and correctness under hardware-specific execution models.

As illustrated in Fig.~\ref{fig:framework}, the dataset consists of three complementary components: documentation-based reasoning data derived from AscendC manuals, kernel-centric chain-of-thought (CoT) data extracted from real-world AscendC operators, and general reasoning chains to preserve generalization capability. In total, the raw corpus contains 83,916 samples and is publicly available at \href{https://huggingface.co/AscendKernelGen/datasets}{\textcolor{blue}{Ascend-CoT}}.

The raw dataset exhibits a long-tailed length distribution. Approximately 99.1\% of samples have input sequences shorter than 11.1k tokens, although the maximum input length reaches about 111k tokens. On the output side, 85.1\% of samples are within 21.9k tokens and an additional 6.3\% fall between 21.9k and 43.8k tokens, while the maximum output length reaches about 219k tokens. Before supervised fine-tuning, we apply length-aware preprocessing: samples exceeding 32k tokens are discarded, and short samples are packed into fixed-length training sequences using boundary-aware isolation to prevent interference across concatenated examples. After preprocessing, the final SFT corpus consists of 9,955 sequences of length 32k.

\subsection{Documentation-Based CoT Data}
\label{document_dataset}

The AscendC programming ecosystem contains rich but highly specialized knowledge spanning hardware abstractions, programming models, and low-level API semantics. To inject this domain knowledge into LLMs in a controllable manner, we transform official AscendC documentation into structured, reasoning-oriented supervision rather than treating it as unstructured text.

Specifically, we construct documentation-based supervision from three authoritative sources: the AscendC Operator Development Guide, which introduces hardware abstractions and the programming model; the AscendC API Reference, which specifies kernel, host, and debugging interfaces; and the AscendC Best Practices, which describe performance optimization strategies and common pitfalls. Together, these sources provide comprehensive coverage of both functional correctness and performance-oriented considerations.

Instead of unsupervised continual pre-training, we adopt a \emph{knowledge-instruct} paradigm that converts documentation content into question–-answer pairs augmented with explicit reasoning traces. The resulting supervision emphasizes not only factual knowledge of APIs and concepts, but also the reasoning process underlying correct API usage and constraint satisfaction. This design mitigates hallucination caused by shallow pattern memorization and encourages systematic reasoning about hardware and programming constraints.

The constructed documentation-based dataset covers both conceptual reasoning over hardware abstractions and programming semantics, as well as API-centric reasoning that focuses on invocation patterns, parameter constraints, and typical error scenarios. Each sample consists of a natural-language question, a structured chain of reasoning, and a grounded answer. To ensure coverage and diversity, we enforce fine-grained chapter alignment and apply paraphrase augmentation during data generation.

\subsection{Code-Centric CoT Data}
\label{kernel_dataset}

Beyond documentation, correct kernel generation requires understanding how host-side tiling logic interacts with device-side kernel execution in practice. To capture this interaction, we construct code-centric chain-of-thought (CoT) data from real-world AscendC operator implementations.

For standalone kernel files that can be compiled independently, we generate kernel-level CoT samples that explicitly explain kernel structure, memory movement, and computation logic. Each sample pairs a functional specification with a complete kernel implementation and an explicit reasoning trace that justifies key design decisions, such as tiling assumptions, buffer usage, and pipeline organization.

However, industrial-grade AscendC operators are typically implemented as composite projects that support multiple shapes, data types, and execution branches. These implementations interleave host-side tiling logic with device-side kernel invocation, making them unsuitable for direct supervision. To address this challenge, we systematically decompose complex operators into logically pure host--kernel pairs, where each pair corresponds to a single execution scenario with fixed tiling and shape assumptions.

For each decomposed host--kernel pair, we construct project-level CoT supervision that jointly reasons about host-side tiling parameter computation, kernel-side pipeline structure and memory staging, and the consistency between tiling metadata and kernel execution behavior. This joint reasoning explicitly exposes the cross-boundary dependencies that are essential for correct kernel generation but are rarely visible in isolated kernel code.

To ensure correctness and executability, we retain only CoT samples that pass automated verification against reference tiling configurations. This filtering step guarantees that the generated reasoning traces are not only conceptually coherent but also aligned with valid host--kernel execution semantics.

\subsection{General CoT Data}
\label{general_cot}

While domain-specific supervision is essential, training exclusively on kernel data risks degrading general reasoning ability. To balance specialization and generality, we augment our dataset with high-quality open-source chain-of-thought corpora covering mathematics, code reasoning, and scientific problem solving~\cite{Chinese-Data-Distill-From-R1,AM-DeepSeek-R1-0528-Distilled,r1-dataset-collection-2025,reasoning_gemini_300k_2025}.

We apply a unified filtering and normalization pipeline to ensure consistency across heterogeneous sources, including language normalization, length-based filtering, and perplexity-based quality control. This mixture allows the model to retain general problem-solving skills while adapting to the highly specialized reasoning demands of NPU kernel generation.

\section{LLM Fine-Tuning for NPU Kernels}\label{fine_tune}


To enable reliable NPU kernel generation under strict programming constraints, we introduce \href{https://huggingface.co/AscendKernelGen/models}{\textcolor{blue}{KernelGen-LM}}, a publicly available domain-adapted model designed to internalize AscendC-specific API semantics and complex execution dependencies. Our approach employs a two-stage optimization strategy combining Supervised Fine-Tuning (SFT) and Reinforcement Learning (RL). The first stage performs NPU-aware supervised fine-tuning (SFT) with curated and iteratively refined data, enabling the model to acquire foundational knowledge of AscendC syntax, APIs, and kernel structure. The second stage further refines the model via reinforcement learning (RL) using execution-based correctness signals, encouraging the generation of kernels that are not only compilable but also numerically correct.

\subsection{Supervised Fine-Tuning with Error-Derived Supervision}

\begin{wrapfigure}{l}{0.48\linewidth}
    \centering
    \includegraphics[width=\linewidth]{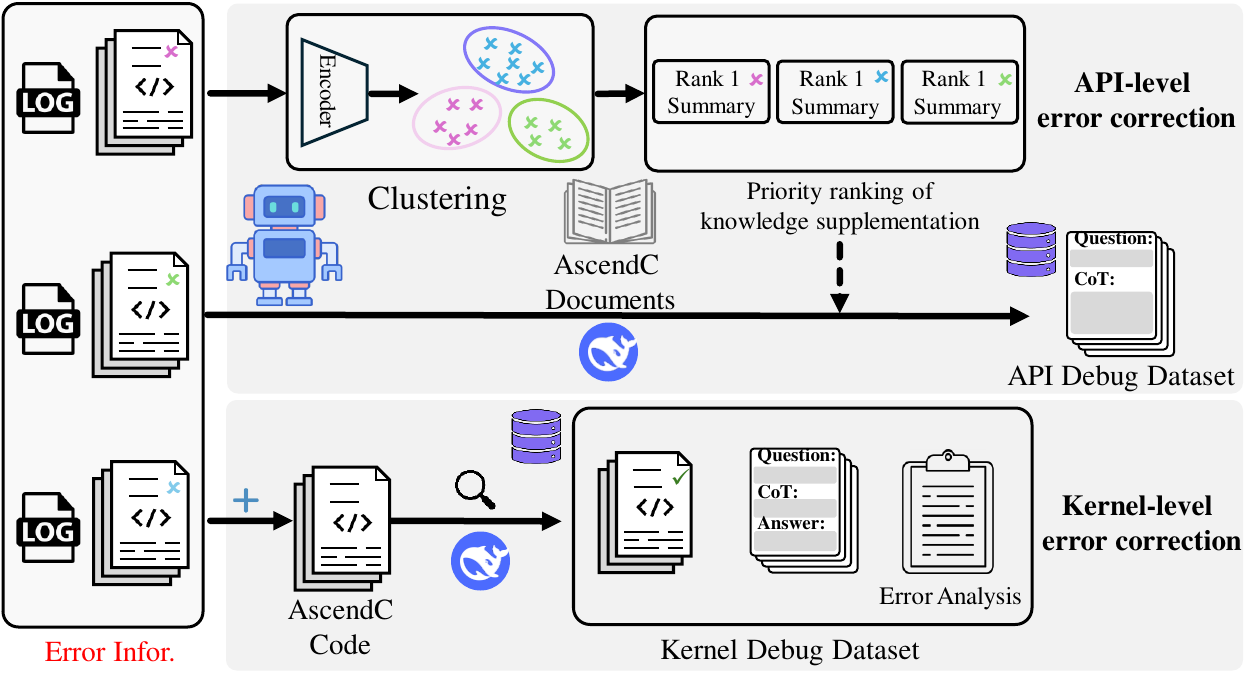}
    \caption{Error-Derived Supervision for API- and Kernel-Level Error Correction.}
    \vspace{-10pt}
    \label{fig:error_correction}
\end{wrapfigure}

While Ascend-aware SFT provides essential syntactic and structural knowledge, we observe that models trained solely on static supervision still exhibit frequent failures in compilation stability and numerical correctness. To improve robustness prior to RL, we introduce an \emph{error-derived supervision} strategy that augments SFT with correction data generated from real execution failures, shown in Figure~\ref{fig:error_correction}. 
This strategy focuses on two complementary levels of errors: API-level misuse and kernel-level numerical inconsistency.

\subsubsection{API-Level Error Correction}
\label{sft_api_correction}

At the API level, we construct an error-correction dataset from real AscendC compilation failures. Given a compiler error log, the corresponding kernel context, and relevant official documentation, the model is trained to identify the root cause of the failure and produce a corrected implementation. This supervision explicitly targets a common failure mode of LLM-based kernel generation, where API invocations are syntactically plausible but violate semantic constraints or usage requirements.

The overall construction process is summarized in Algorithm~\ref{alg:api_sft_pipeline}, which formalizes how compilation errors are analyzed and transformed into supervised correction examples.

\begin{algorithm}[t]
\caption{Workflow for Constructing Ascend C API Correction SFT Data}
\label{alg:api_sft_pipeline}

\KwIn{
  Raw build logs; \\
  Source-code repository; \\
  OCR-extracted API documentation
}
\KwOut{SFT dataset with reasoning and corrected API usage}

\textbf{Step 1: Extract Error Logs} \\
\Begin{
  Traverse all build logs; \\
  \ForEach{log file}{
    \If{contains error messages}{
      Extract the error block and save it as a clean error sample;
    }
  }
}

\vspace{1mm}

\textbf{Step 2: Error Distribution Analysis} \\
\Begin{
  Classify error types using an LLM; \\
  Compute pass/fail statistics to identify challenging operators; \\
  Optionally filter easy cases using the statistics.
}

\vspace{1mm}

\textbf{Step 3: Contextual Diagnostic Report Generation} \\
\Begin{
  \ForEach{error sample}{
    Retrieve related code snippet and variable definitions; \\
    Retrieve the corresponding API documentation; \\
    Merge log excerpt, code context, and API reference into a structured Markdown report;
  }
}

\vspace{1mm}

\textbf{Step 4: SFT Data Synthesis} \\
\Begin{
  \ForEach{diagnostic report}{
    Prompt an LLM to: \\
    \Indp analyze the misuse of the API; \\
    produce expert reasoning (\texttt{<think>}); \\
    propose corrected usage and explanations; \Indm \\
    Save the result as an SFT training instance;
  }

  Apply data balancing via subsampling or augmentation; \\
  Remove low-quality or incomplete samples.
}

\end{algorithm}

\subsubsection{Kernel-Level Error Correction}\label{sft_kernel_correction}


Beyond compilation errors, a substantial fraction of generated kernels compile successfully but fail numerical verification. Such failures typically arise from subtle kernel-level reasoning issues, including incorrect memory staging, accumulation order mismatches, or inconsistencies between host-side tiling metadata and device-side execution logic.

To address these errors within the supervised learning paradigm, we construct ground-truth–-guided reconstruction data. As formalized in Algorithm~\ref{alg:precision_sft_pipeline}, kernels that pass compilation but fail numerical verification are first identified and then paired with their corresponding ground-truth (GT) implementations, which serve as deterministic supervision targets.

Based on these pairings, we generate reconstruction-oriented chain-of-thought supervision that guides the model to analyze the causes of numerical failures and synthesize corrected kernel implementations from scratch. Unlike reinforcement learning, this process does not rely on scalar rewards or preference comparisons, but instead enforces hard execution and numerical invariants through explicit example-level correction.

The resulting correction-derived samples are incorporated into the SFT corpus, enabling the model to internalize numerically sensitive execution constraints and significantly reducing the prevalence of silent correctness failures. By eliminating a large portion of invalid candidates prior to reinforcement learning, this error-derived SFT stage stabilizes subsequent policy optimization and improves sample efficiency by narrowing the search space to reliably executable kernels.

\begin{algorithm}[t]
\caption{Pipeline for constructing precision-corrected SFT data}
\label{alg:precision_sft_pipeline}

\KwIn{Collection of test logs; repository of GT implementations}
\KwOut{SFT dataset with reasoning and corrected kernels}

\textbf{Step 1: Identify Precision-Failure Samples} \\
\Begin{
  Traverse all log directories; \\
  \ForEach{sample}{
    \If{compilation succeeds \textbf{and} precision check fails}{
      Extract its corresponding JSON metadata;
    }
  }
}

\vspace{1mm}

\textbf{Step 2: Generate Reconstruction-Based Training Data} \\
\Begin{
  \ForEach{extracted sample}{
    Remove existing reasoning traces; \\
    Load erroneous kernel code; \\
    Retrieve corresponding GT kernel implementation; \\

    Build an LLM prompt combining: \\
    \Indp problem description, erroneous code, and GT reference; \Indm \\

    Invoke LLM to: \\
    \Indp
      analyze root cause of precision failure; \\
      discard faulty implementation; \\
      reconstruct a correct kernel with new \texttt{<think>} and \texttt{<kernel\_impl>}; 
    \Indm \\

    Store the generated result as an SFT instance;
  }
  Perform resampling to satisfy the desired data quota per kernel;
}

\end{algorithm}

\subsection{Reinforcement Learning with Execution-Based Preferences}\label{reinforcement_learning}



After error-derived supervised fine-tuning, the model is able to generate kernels that largely satisfy compilation constraints and basic numerical correctness. However, multiple candidate kernels may still exist for a given specification, differing in subtle execution behaviors such as memory access patterns, accumulation order, or numerical stability. These distinctions are difficult to encode through deterministic supervision alone.

To further refine the model, we introduce a reinforcement learning (RL) stage driven by execution-based preference signals. Unlike the preceding SFT stage, which focuses on correcting explicit failures, the RL stage operates over kernels that already pass compilation and verification, and aims to preferentially reinforce higher-quality implementations.

As shown in Fig.~\ref{fig:framework}, for each kernel generation task, the model samples multiple candidate implementations, all of which are executable. These candidates are evaluated using an automatic execution-based verifier, which produces preference relations based on numerical accuracy and execution correctness. Rather than constructing explicit correction targets, we form preference pairs: kernels that pass both compilation and precision tests are treated as positive samples, while kernels that compile but fail precision tests are treated as negative samples. These pairs define a relative preference signal indicating which generations are more desirable. The resulting preferences are then used to update the model, encouraging behaviors that consistently lead to more stable and accurate kernel implementations.

This RL stage benefits from the preceding error-derived SFT process. By eliminating a large fraction of invalid or numerically incorrect candidates beforehand, supervised correction significantly narrows the policy search space and stabilizes preference learning. As a result, reinforcement learning focuses on fine-grained optimization among valid kernels rather than recovering from catastrophic execution failures, leading to improved training efficiency and convergence stability.

\section{The Evaluation Sub-system: NPUKernelBench}\label{NPUKernelBench}

\subsection{Benchmark Overview}

Figure~\ref{fig:framework} includes an overview of \href{https://github.com/weich97/NPUKernelBench}{\textcolor{blue}{NPUKernelBench}}, an end-to-end evaluation framework for assessing the ability of large language models (LLMs) to generate correct and efficient NPU kernels. The framework spans the full evaluation lifecycle, from task specification and code generation to compilation, hardware execution, and quantitative measurement.

Given a structured kernel task and corresponding instructions, NPUKernelBench prompts the LLM to generate both a structured chain-of-thought (CoT) representation and executable host-side and kernel-side code. The generated code is then compiled and executed on the NPU hardware, where functional correctness and execution performance are automatically evaluated against reference implementations.

By grounding evaluation in actual hardware behavior and enforcing full compilation and execution, NPUKernelBench enables objective, reproducible, and fine-grained analysis of kernel generation quality.

\subsection{Kernel Task: Hierarchical and Categorical Design}


Evaluating LLMs for NPU kernel generation requires disentangling multiple sources of difficulty. Treating all kernels as a homogeneous benchmark obscures key variations in computational structure and interface requirements, leading to incomplete assessments of model capability.

NPUKernelBench addresses this challenge through a structured task design that decomposes kernel difficulty along two orthogonal dimensions: \emph{algorithmic complexity} and \emph{interface complexity}. Together, these dimensions form a hierarchical and categorical evaluation scheme that enables systematic analysis of both specialization and generalization behavior.

\paragraph{Algorithmic Complexity.}
Kernel tasks are categorized into three levels according to their inherent computational structure (Table~\ref{tab:kernels}). Level~1 tasks consist of simple element-wise or arithmetic operations with linear data flow. Level~2 tasks correspond to common neural network operators with structured computation and local dependencies. Level~3 tasks include operators with global dependencies, iterative computation, or dynamic control flow, such as \texttt{Gemm}, \texttt{TopK}, and attention-related kernels. These tasks place increasing demands on parallel reasoning, memory hierarchy management, and control logic synthesis.
\begin{table*}[ht]
\centering
\caption{Categorization of kernel tasks in NPUKernelBench.}
\resizebox{1\linewidth}{!}{
\begin{tabular}{clllll}
\toprule
\textbf{Level} & \textbf{Static-shape} & \textbf{Dynamic-shape} & \textbf{Categories} & \textbf{Representative} & \textbf{Tasks} \\
\midrule
\multirow{5}{*}{Level 1} &\multirow{5}{5.5cm}{Can linear data flow be mapped to fixed-size loops with deterministic memory access?}&\multirow{5}{5.5cm}{Can element-wise logic remain correct under runtime-determined tensor shapes?} & Comparison & Add, Equal & 3 \\
&&&Condition & IsFinite, IsInf & 3 \\
&&&Index & GatherV3, ScatterList & 2 \\
&&&Math &AddCustom, Sqrt & 26 \\
&&&TensorCreation &Arange, Eye &3 \\ \hline
\multirow{9}{*}{Level 2} &\multirow{9}{5.5cm}{Can structured computation exploit fixed tiling and local data reuse?}& \multirow{9}{5.5cm}{Can kernels generalize correctly across varying input sizes?} & Activation & Gelu, MulSigmoid & 14 \\
&&&Foreach & ForeachAbs, ForeachSqrt& 53\\
&&&Linalg & Cos, Matmul &2 \\
&&&Loss & CrossEntropyLoss, MseLoss & 4\\
&&&Mask & Tril, Triu & 3\\
&&&Norm & AddLayerNorm, RmsNorm &16\\
&&&Optim &ApplyAdamWV2& 2\\
&&&Reduce & MulSigmoidMulAddCustom & 4\\
&&&TensorMove & ExpandV2, ReverseSequence & 5\\ \hline
\multirow{4}{*}{Level 3} &\multirow{4}{6cm}{Can complex kernels with global dependencies and non-trivial execution logic be generated correctly?}& \multirow{4}{5.5cm}{Can kernels support dynamic shapes and control flow in complex execution scenarios?} & \multirow{2}{*}{GMM} & \multirow{2}{*}{Gemm, BasicMatmul} & \multirow{2}{*}{17} \\
&&& \multirow{2}{*}{Sort} & \multirow{2}{*}{TopKV3} & \multirow{2}{*}{1} \\
&&&&&\\
&&&&&\\
\midrule
&&&Total & 16 & 158 \\
\bottomrule
\end{tabular}}
\label{tab:kernels}
\end{table*}

\vspace{-1mm}
\begin{wrapfigure}{r}{0.48\linewidth}
    \centering
    \vspace{-10pt}
    \includegraphics[
  width=\linewidth,
  trim=230pt 50pt 250pt 80pt,
  clip
]{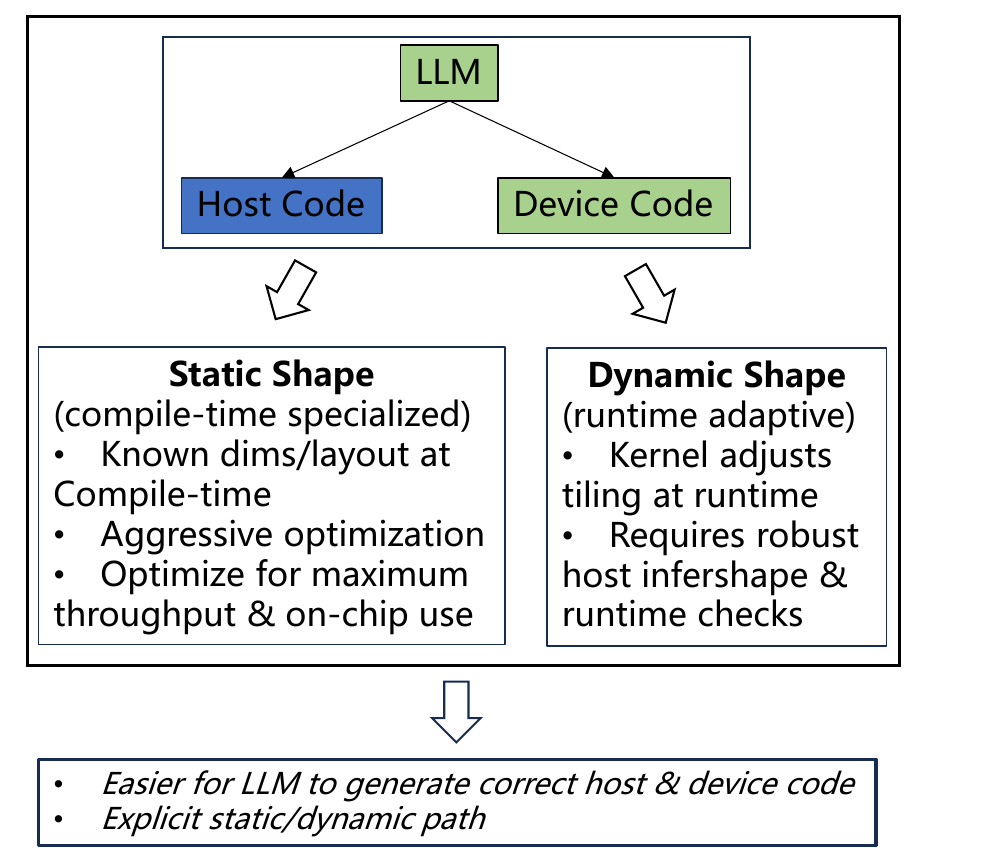}
    \vspace{-10pt}
    \caption{Dual-Path Evaluation Design.}
    \label{fig:dual_path}
    \vspace{-20pt}
\end{wrapfigure}
\paragraph{Interface Complexity.}
In addition to algorithmic difficulty, NPUKernelBench distinguishes between static-shape and dynamic-shape kernel tasks, shown in Figure~\ref{fig:dual_path}. Static-shape tasks assume compile-time known tensor dimensions and primarily evaluate specialization and optimization capability. Dynamic-shape tasks require kernels to adapt to runtime-determined dimensions, testing the model’s ability to generate robust host-side logic for shape inference, tiling computation, and control flow.

\subsection{Standardized Generation Interface}\label{sec_generate_kernel}

To ensure fair and reproducible evaluation, NPUKernelBench defines a standardized generation interface that constrains both the input prompt and the structure of the generated code. This design bridges the gap between probabilistic LLM outputs and the deterministic requirements of the NPU compilation and execution pipeline.


\subsubsection{Prompt Formulation}
As illustrated in Tables~\ref{tab:prompt_ins} and \ref{tab:kernel_gen}, the prompt provided to the LLM integrates three complementary components: (1) task specification via API descriptions, (2) structural scaffolding through host and kernel code templates, and (3) role-based and formatting constraints. Together, these components provide sufficient semantic grounding and structural guidance for generating compilable Ascend C code.

\textbf{Task Specification via API Description.} The core functional requirements are encapsulated in \texttt{API\_Desc.md}. As shown in Table~\ref{tab:kernel_gen}, this module translates the mathematical definition of the kernel into machine-readable specifications, detailing input/output tensor attributes (shapes, data types, layouts) and hardware constraints. For example, for a matrix multiplication kernel, the prompt explicitly formalizes the computation rule $C_{ij} = \sum_{p=1}^{k} A_{ip} B_{pj}$ and specifies precision modes, ensuring the model grounds its generation in accurate arithmetic logic.

\textbf{Structural Scaffolding via Code Templates.} To decouple algorithmic logic from boilerplate syntax, NPUKernelBench injects pre-defined code templates located in the \texttt{tasks/.../question/} directory. The prompt includes a host template (defining runtime registration and tiling interfaces) and a kernel template (specifying device-side entry points, e.g., \texttt{basic\_matmul\_kernel}). Table \ref{tab:kernel_gen} in the Appendix demonstrates a full template instance. This scaffolding forces the LLM to complete valid C++ structures rather than generating free-form text, significantly improving the compilability of the output.
\begin{wraptable}{l}{0.5\textwidth}
\vspace{-10pt}
\centering
\begin{tcolorbox}[
      sharp corners,
      colframe=reforange,
      colback=reforange!5,
      colbacktitle=reforange!15,
      coltitle=reforange,
      boxrule=0.4pt,
      title={\textbf{Instructions}},
      fonttitle=\bfseries\small,
      left=0.5pt,
      right=0.5pt,
      top=0.5pt,
      bottom=0.5pt
    ]
{\small
\begin{itemize}[leftmargin=*, itemsep=1pt, parsep=0pt, topsep=2pt]
    \item \textbf{Role:} "You are a professional Ascend kernel development engineer."
    \item \textbf{Target Language and API:} "Please use Ascend C for development."
    \item \textbf{Code Style Requirements:} "Ensure high code readability and include necessary comments."
    \item \textbf{Output Format Requirements:} "Generate separate Kernel and Host code: the Kernel defines <kernel\_name>\_kernel, and the Host declares Operator<OperatorName>Paras and implements the corresponding computation logic."
    \item \textbf{Key Information Reminder:} "Pay attention to handling the tensor data types, shapes, and layouts properly."
\end{itemize}
}
\end{tcolorbox}
\vspace{-10pt}
\captionof{table}{General instructions used in the prompt.}
\label{tab:prompt_ins}
\vspace{-10pt}
\end{wraptable}

At the beginning of a prompt, there are usually several general instructions, as illustrated in the left Table~\ref{tab:prompt_ins}. These instructions provide meta-level guidance that defines the model’s behavior, coding conventions, and output formatting. Specifically, they include explicit role prompts (e.g., adopting the perspective of a professional Ascend kernel engineer), target language and environment specifications (e.g., using Ascend C), code style requirements (e.g., ensuring readability and including necessary comments), and output format constraints (e.g., generating separate Kernel and Host code with clearly defined function and structure names). By providing sufficiently detailed instructions, the framework ensures that the LLM can follow consistent generation patterns and produce code that aligns with the target development standards.

\textbf{Role-Based Constraints.} The prompt begins with meta-instructions (Table~\ref{tab:prompt_ins}) that configure the model's persona (e.g., "Professional Ascend Kernel Development Engineer") and impose strict formatting rules. These instructions explicitly forbid conversational filler and mandate the separation of Host and Kernel code blocks. By formalizing these constraints, the framework ensures that the raw output from the LLM requires minimal post-processing before entering the compilation pipeline.


Overall, the prompt formulation follows a deterministic sequence: ingesting the \texttt{api\_desc.md}, appending structural templates, and prepending constraint instructions. This creates a cohesive input context that maximizes the model's potential for generating compliant Ascend C code.

\subsubsection{Output Code Specification \& Dual-Path Evaluation}

NPUKernelBench enforces a strict output code specification to enable fully automated compilation and execution. Based on this specification, the framework supports a dual-path evaluation mechanism that assesses kernel generation capability under different system assumptions.

Specifically, NPUKernelBench supports two complementary evaluation paths. In the \emph{Device-Only Path}, the framework provides a fixed, pre-optimized host driver and evaluates the LLM solely on its ability to generate correct and efficient kernel-side code. This setting isolates the model’s fundamental capability in implementing core computational logic on the NPU and is particularly suitable for L1/L2 operators with minimal control flow. In contrast, the \emph{Host+Device Path} evaluates the model’s ability to generate complete, deployable operators. Under this setting, the LLM is required to produce both host-side control logic and kernel-side implementation, reflecting realistic deployment scenarios where runtime scheduling, shape inference, and dynamic memory management are essential.

To support the Host+Device path, the output specification requires the generated code to instantiate three logically distinct but tightly coupled components within the provided templates (Table~\ref{tab:kernel_gen}). First, the host-side logic executes on the CPU and is responsible for orchestrating operator execution. This includes defining parameter structures for host–device communication, performing shape inference to derive output tensor dimensions at runtime, computing tiling parameters tailored to the target NPU architecture, and configuring kernel launch dimensions. Second, the kernel-side logic executes on the NPU AI Core and implements the parallel computation itself. The generated kernel must conform to the Ascend C programming model, correctly manage on-chip memory hierarchies, synchronize parallel execution, and employ SIMD-style computation with techniques such as double buffering to overlap data movement and computation. Finally, for operators with decoupled tiling logic, the framework expects an explicit tiling structure definition that bridges host-side tiling computation and kernel-side execution.

This dual-path design enables NPUKernelBench to comprehensively evaluate kernel generation capability across two representative industrial scenarios, shown in Figure~\ref{fig:dual_path}. For \emph{static shape optimization}, the framework analyzes the generated kernel and tiling logic to assess whether the model can produce highly optimized code for fixed problem sizes, focusing on tiling strategy design, efficient use of on-chip memory, and compute–memory pipeline optimization. For \emph{dynamic shape robustness}, enabled by the Host+Device path, the framework evaluates the model’s ability to generalize to runtime-varying tensor shapes. In this setting, correctness and performance depend on adaptive host-side logic—such as dynamically computed tiling parameters and loop bounds—thereby testing the robustness of LLM-generated code in realistic deployment environments.


By standardizing both prompt formulation and output structure, NPUKernelBench isolates model capability from prompt engineering artifacts and ensures that evaluation outcomes reflect genuine differences in kernel generation ability rather than formatting or interface inconsistencies.

\subsection{Automated Evaluation Pipeline}

NPUKernelBench is not merely a collection of test cases but a resilient, end-to-end automation framework designed to bridge the gap between model generation and rigorous hardware verification. As illustrated in Fig.~\ref{fig:framework}, the pipeline seamlessly integrates four traditionally isolated stages: \textit{Code Generation}, \textit{Compilation}, \textit{Correctness Verification}, and \textit{Performance Benchmarking}.

\subsubsection{End-to-End Architecture and Concurrency}


To support large-scale and reproducible experiments (e.g., $N \geq 100$ samples per task), the framework follows a \emph{one-click} execution philosophy. The entire evaluation process is initiated from a single entry point (\texttt{run\_multi\_test.py}) and configured through a unified definition file (\texttt{base\_config.yaml}), eliminating manual intervention and reducing inconsistencies across experimental runs.

At scale, evaluating thousands of LLM-generated kernels across diverse NPU operators requires not only parallelism but also strong fault isolation. To this end, NPUKernelBench adopts a high-concurrency execution model based on Python \texttt{multiprocessing}, enabling parallel compilation and execution to maximize host CPU and NPU utilization. Each kernel evaluation is executed within a strictly isolated subprocess, ensuring that fatal errors—such as segmentation faults, illegal memory access, or NPU runtime exceptions—are contained locally and never propagate to disrupt the global pipeline.

Given the stochastic nature of LLM outputs and the complexity of hardware execution, the pipeline further incorporates native stability mechanisms to guarantee long-running robustness. Specifically, timeout control is used to proactively terminate non-terminating kernels or hardware hangs, while an automatic retry mechanism mitigates transient failures caused by network instability during API calls or temporary resource contention. In parallel, an active resource monitor enforces strict cleanup of host memory and NPU VRAM after each subprocess terminates, preventing resource leakage and ensuring consistent system behavior throughout extended batch evaluations.

Beyond execution robustness, NPUKernelBench is designed as a dedicated benchmarking suite with a structured, multi-level evaluation methodology tailored to custom NPU kernels. The framework evaluates generated operators along two primary dimensions: correctness and utilization, while adhering to three guiding principles. First, \emph{Correctness First} establishes functional validity as a prerequisite for any performance assessment. Second, \emph{Performance-Oriented Evaluation} emphasizes the effective utilization of NPU hardware resources once correctness is ensured. Third, \emph{Multi-Level Difficulty} introduces progressively more challenging tasks, enabling systematic assessment of optimization capability from basic implementations to highly tuned kernels.

These principles are realized through a fully automated evaluation pipeline comprising three tightly integrated stages: compilation evaluation, correctness evaluation, and performance evaluation. Generated kernels are first compiled to verify syntactic and build correctness, then executed on available NPU devices to validate numerical accuracy, and finally benchmarked to collect performance metrics such as execution latency and resource utilization. All stages are scheduled and executed in parallel whenever possible, with detailed logs and intermediate results recorded automatically. This design enables large-scale, objective, and reproducible evaluation of kernel generation quality with minimal human intervention.

\subsubsection{Compilation Evaluation}
In this stage, each generated kernel in Sec.~\ref{sec_generate_kernel} is first verified to ensure that it conforms to the Ascend C syntax and can be successfully compiled. To achieve this, the kernel is processed using the Ascend C compiler, and the compilation status is recorded as either success or failure. Meanwhile, detailed logs are stored in the \texttt{log/} directory of each kernel sample, which allows developers to inspect the corresponding \texttt{log\_file} in order to identify and diagnose any compilation errors. To illustrate this process, Table~\ref{tab:comp_log} presents a sample compilation log, highlighting the successful build of the \texttt{Sqrt} kernel from Level 1.

\begin{table}[H]
\vspace{-10pt}
\centering
\begin{tcolorbox}[
      sharp corners,
      colframe=refpurple,
      colback=refpurple!5,
      colbacktitle=refpurple!15,
      coltitle=refpurple,
      boxrule=0.4pt,
      title={\textbf{Compilation Log}},
      fonttitle=\bfseries\small,
      left=0.5pt,
      right=0.5pt,
      top=0.5pt,
      bottom=0.5pt
    ]
    \begin{lstlisting}[style=mystyle]
[INFO] Task lvl1_categoryMath_Sqrt_sample0 compile result:
==========================================================================================================
CompileResult:
success = True
log_file:
runs/msopgen/lvl1/Math/Sqrt/fixed_case_0/sample0/log/lvl1_categoryMath_Sqrt_sample0_compile.log
==========================================================================================================
\end{lstlisting}
\end{tcolorbox}
\captionof{table}{Compilation log of the \texttt{Sqrt} kernel using the Ascend C compiler. }
\label{tab:comp_log}
\vspace{-10pt}
\end{table}

\subsubsection{Correctness Evaluation}\label{kernelbench_correctness}


The correctness evaluation stage assesses whether LLM-generated kernels produce numerically valid results across diverse input configurations. For each kernel task, NPUKernelBench defines a set of test cases that collectively cover relevant combinations of data types and tensor shapes. Each generated kernel is automatically compiled and executed, and correctness is determined by element-wise comparison against a pre-generated \textit{golden reference} under predefined numerical tolerances. These test outcomes form the basis of a unified and hierarchical correctness scoring scheme.

At the task level, correctness is quantified using a single \textbf{Task-Level Score}, defined as the proportion of test cases that pass successfully:
\begin{equation}
   Score_{task} = \frac{\text{Number of Passed Test Cases}}{\text{Total Number of Test Cases}} \times 100.
\end{equation}
This abstraction intentionally treats all test cases within a task uniformly, providing a simple yet robust measure of functional validity. A kernel achieves a full score when it passes all test cases, while partial correctness is reflected by a proportionally reduced score. Kernels that fail all test cases receive a score of zero, reinforcing the principle that correctness is a strict prerequisite for any further evaluation.

The same task-level metric is applied consistently to both static-shape and dynamic-shape tasks (Table~\ref{tab:kernels}). For static-shape tasks, all test cases correspond to a single fixed input configuration and jointly verify correctness for that shape. In contrast, dynamic-shape tasks require a single kernel implementation to handle multiple input shapes. As a result, multiple test cases are associated with each task, making shape generality an inherent requirement for achieving a high task-level score. Failure on any required shape directly reduces the score, naturally exposing limitations in robustness and adaptability.

To summarize correctness performance at higher levels of abstraction, NPUKernelBench further defines a \textbf{Difficulty-Level Score}. For each difficulty tier (L1, L2, and L3), this score is computed as the arithmetic mean of all task-level scores within that tier:
\begin{equation}
   Score_{Lk} = \operatorname{Average}_{i \in \text{Lk tasks}} (Score_{task_i}), \quad k \in \{1,2,3\}.
\end{equation}
This aggregation provides a stable estimate of correctness performance across tasks of similar complexity, mitigating variance introduced by individual operators.

Finally, an overall \textbf{Total Correctness Score} is computed as a weighted sum of the difficulty-level scores:
\begin{equation}
   Score_{final} = (Score_{L1} \times w_{L1}) + (Score_{L2} \times w_{L2}) + (Score_{L3} \times w_{L3}),
\end{equation}
where higher weights are assigned to more challenging tasks to emphasize correctness on complex operators. By default, the framework uses $w_{\text{L1}} = 0.2$, $w_{\text{L2}} = 0.3$, and $w_{\text{L3}} = 0.5$. This design encourages models to prioritize correctness on difficult tasks and serves as a gating signal for subsequent performance evaluation. Scoring criteria for kernel evaluation is given in Table~\ref{tab:level_correctness}. Representative evaluation criteria under both default and customized settings are also given in Table~\ref{tab:kernel_correctness}.

\begin{table}[t]
\centering
\caption{Scoring criteria for kernel evaluation.}
\resizebox{1\linewidth}{!}{
\begin{tabular}{lll}
\hline
\textbf{Scoring Item} & \textbf{Formula} & \textbf{Core Idea} \\
\hline
Task Score & $ \text{Score}_{\text{task}} = \frac{\text{Number of Passed Test Cases}}{\text{Total Number of Test Cases}} \times 100 $ & Measures correctness and serves as the foundation of the overall score. \\
Level Score & $ \text{Average}(\text{Score}_{\text{task}_j}) $ & Reflects the overall performance at a given difficulty level. \\
Total Score & $ \sum (\text{Score}_{\text{level}} \times w_{\text{level}}) $ & Weighted sum incentivizing completion of higher-difficulty tasks. \\
\hline
\end{tabular}}
\label{tab:level_correctness}
\end{table}

\subsubsection{Performance Evaluation}

The performance evaluation stage measures the execution efficiency of LLM-generated kernels on the target NPU. The primary metric is kernel execution latency. To ensure stability and comparability, each test case is executed with a fixed number of warm-up iterations to mitigate cold-start effects, followed by $N$ consecutive runs. The average execution time across these runs is reported as the performance measurement for the test case.

Rather than reporting absolute latency alone, NPUKernelBench evaluates performance relative to a hardware-aware reference upper bound, denoted as $T_{\text{ref}}$. This reference time reflects the execution latency achieved by highly optimized vendor-provided kernels for the same operator and tensor configuration on the target NPU. $T_{\text{ref}}$ does not represent a strict theoretical lower bound on execution time. Instead, it serves as a strong and practically attainable performance reference, instantiated using mature and well-engineered implementations from official NPU libraries (e.g., \texttt{aclnn}). While such vendor kernels may not always achieve the absolute hardware limits, they provide a stable, reproducible, and hardware-aware baseline that reflects current best engineering practices.

By normalizing kernel latency against this reference, the evaluation emphasizes how effectively a generated kernel approaches the performance level of optimized production kernels, rather than relying on raw execution time alone. This relative formulation mitigates sensitivity to operator-specific scale, tensor shapes, and hardware characteristics, enabling fair comparison across heterogeneous workloads. As in the correctness evaluation, performance is measured only for kernels that pass functional validation, ensuring that efficiency is assessed exclusively for correct implementations.

\subsection{Optimization-Ready Fine-Grained Logging}
\label{sec:logging_system}

The logging system of NPUKernelBench is designed not merely for passive record-keeping, but as a feedback mechanism that bridges evaluation and model optimization. By systematically capturing detailed execution traces throughout the evaluation pipeline, the framework provides fine-grained, actionable signals that support iterative improvement of LLM-based code generation models.

At each stage of the pipeline, including code generation, compilation, and correctness verification, NPUKernelBench produces structured and stage-aware logs. Rather than reporting only coarse-grained outcomes (e.g., a binary build failure), the framework records precise diagnostic information such as compiler error codes, line numbers, and mismatch details. These logs are stored in a structured format (e.g., \texttt{compile.log}), ensuring that every failure mode is traceable, reproducible, and amenable to downstream analysis.

This design transforms evaluation outputs from sparse scalar signals into rich feedback representations. Instead of reducing kernel quality to a single pass/fail outcome, the logging system exposes intermediate failure modes and partial successes, enabling more informative supervision. Such feedback can directly support manual prompt refinement as well as automated optimization pipelines, including supervised fine-tuning (SFT), where structured error patterns can be leveraged as learning signals.

Beyond supervised settings, the fine-grained logging mechanism further enables reinforcement learning based optimization. By decomposing kernel generation into a sequence of verifiable milestones, such as syntactic validity, successful compilation, numerical correctness, and performance attainment, the framework can provide dense and interpretable reward signals. These signals can be naturally mapped to reward shaping strategies, allowing incremental credit assignment during training. As a result, NPUKernelBench extends beyond a static benchmarking suite and can function as an evaluation-driven training environment, supporting the development of self-improving kernel generation agents.

\section{Experiments and Results}
\label{experiments}

\subsection{Experimental Setups}

\textbf{Models.}
As summarized in Table~\ref{tab:models_training}, we employ different large language models at distinct stages of the pipeline according to their functional roles. For dataset construction and document-level reasoning, we adopt DeepSeek-R1~\cite{abs250112948}, which is used to generate document-based data and single-file Chain-of-Thought (Secs.~\ref{document_dataset} and~\ref{kernel_dataset}). For project-level reasoning that requires understanding multi-file kernel implementations with cross-file dependencies, we employ Gemini~2.5~Pro~\cite{google_gemini_2_5_pro}, leveraging its stronger support for long-context and multi-file code reasoning (Sec.~\ref{kernel_dataset}). Within the supervised fine-tuning (SFT) data, DeepSeek-Reasoner is further used to iteratively refine and correct reasoning traces (Secs.~\ref{sft_api_correction} and~\ref{sft_kernel_correction}).

\textbf{Training.}
We adopt a two-stage training strategy consisting of supervised fine-tuning followed by reinforcement learning. In the SFT stage, Qwen3-32B~\cite{qwen3technicalreport} serves as the primary backbone model. To analyze the impact of model scale and architecture, we additionally conduct SFT on multiple Qwen variants ranging from 1.7B to 32B parameters, as well as the code-specialized Qwen3-Coder-30B-A3B-Instruct. All SFT experiments share a unified training configuration, with both full fine-tuning and LoRA-based tuning evaluated under controlled settings (Table~\ref{tab:models_training}). In the second stage, reinforcement learning is performed using Direct Preference Optimization (DPO)~\cite{RafailovSMMEF23}, with detailed configurations and ablation studies reported in Sec.~\ref{reinforcement_learning}.

\begin{table}[t]
\centering
\caption{Overview of models and configurations used for dataset construction and two-stage training.}
\label{tab:models_training}
\resizebox{0.9\linewidth}{!}{
\begin{tabular}{p{3.7cm} p{3.9cm} p{3.9cm} p{5.5cm}}
\toprule
\textbf{Stage} & \textbf{Model / Method} & \textbf{Role} & \textbf{Description} \\
\midrule

\multirow{9}{*}{Dataset Construction} 
& DeepSeek-R1
& Document \& Single-file CoT 
& Document-based data generation and single-file Chain-of-Thought construction \\

& Gemini 2.5 Pro
& Project-level CoT 
& Reasoning over complete kernel projects with multiple mutually referenced source files \\

& DeepSeek-Reasoner 
& Data Flywheel Refinement
& Iterative correction and refinement of generated reasoning chains within the data flywheel \\
\midrule

\multirow{10}{*}{Supervised Fine-tuning}  
& Qwen3-32B
& Backbone Model 
& Backbone model for supervised fine-tuning \\

& Qwen3-1.7B / 4B / 8B / 14B / 32B / Coder-30B-A3B-Instruct 
& Scaling Study 
& Supervised fine-tuning across different model scales and architectures \\

& Training Configuration 
& Full FT / LoRA 
& Learning rate $1.25\times10^{-6}$ (Full FT) vs. $1.25\times10^{-5}$ (LoRA), cosine decay, warm-up ratio 0.01, micro/global batch size 1/128, weight decay $1\times10^{-1}$, gradient clipping 1.0. \\

\midrule

\multirow{7}{*}{Reinforcement Learning}  
& DPO
& Optimization Objective 
& Direct Preference Optimization with $\beta=0.1$ \\

& Training Configuration 
& Hyperparameters 
& Learning rate $1\times10^{-6}$, cosine decay, warm-up ratio 0.1, global batch size 64, 150 training iterations \\

& Preference Construction 
& Ablation Analysis 
& Analysis of different positive and negative sample selection strategies \\
\bottomrule
\end{tabular}}
\end{table}

\subsection{NPUKernelBench and Evaluation Metrics} 

All evaluations are conducted on NPUKernelBench (Sec.~\ref{NPUKernelBench}), which contains 158 representative kernels spanning three levels of computational complexity.
Each kernel is accompanied by a reference implementation with standardized input and output specifications, enabling automated correctness verification.

For each task, we generate kernel code using the evaluated LLM and execute the compiled kernels on the target NPU platform. To assess robustness and scalability, each kernel is evaluated under multiple input sizes, allowing us to examine generalization behavior across varying workload intensities.

We evaluate generated kernels along three complementary dimensions. First, the \emph{Compilation Rate} measures whether a kernel can be successfully compiled, reflecting syntactic validity and API compliance; this metric is reported as pass@$k$, indicating the fraction of tasks for which at least one of the top-$k$ generated candidates compiles successfully. Second, the \emph{Execution Rate} assesses functional correctness by comparing kernel outputs against reference implementations under predefined numerical tolerances (Sec.~\ref{kernelbench_correctness}), and is likewise measured under the pass@$k$ protocol. Finally, \emph{Performance Speedup} evaluates runtime efficiency by normalizing kernel latency against expert-implemented baselines, with values greater than one indicating performance improvements.

\subsection{Main Results} 

We compare kernel generation performance across different training stages, including the base models, supervised fine-tuning, and reinforcement learning. Overall, performance improves consistently as training progresses, with increasingly robust behavior observed on more challenging kernels. Quantitative results under different sampling budgets are summarized in Table~\ref{tab:pass_k_results_speed} and analyzed in detail below.

\begin{table}[t]
\centering
\small
\renewcommand{\arraystretch}{1.1}
\caption{Evaluation results of kernel generation on NPUKernelBench across different sampling budgets ($k$). We report Compilation Rate (CR), Execution Rate (ER), and overall Speedup for generated kernels.}
\label{tab:pass_k_results_speed}
\resizebox{1\linewidth}{!}{
\begin{tabular}{l c cc cc cc c}
\toprule
\multirow{2}{*}{\textbf{Model}} & \multirow{2}{*}{\textbf{Level}} & \multicolumn{2}{c}{\textbf{Pass@1}} & \multicolumn{2}{c}{\textbf{Pass@10}} & \multicolumn{2}{c}{\textbf{Pass@100}} & \multirow{2}{*}{\textbf{Speedup ($\times$)}} \\
\cmidrule(r){3-4} \cmidrule(r){5-6} \cmidrule(r){7-8}
 & & CR (\%) & ER (\%) & CR (\%) & ER (\%) & CR (\%) & ER (\%) &  \\
\midrule
\multirow{4}{*}{Qwen3-32B} & Level 1 & 38.08 & 17.39 &71.62  &48.84  &75.00  &66.67  & 0.60 \\
 & Level 2 & 0.0 & 0.0 & 0.0 & 0.0 & 0.0 & 0.0 & 0.0 \\
 & Level 3 & 1.83 & 0.0 & 15.43 & 0.0 & 50.00 & 0.0 & 0.0 \\
 \rowcolor{bggreen}
 & Mean & 25.59 &11.59  & 49.46 & 32.56 & 55.56 & 44.44 & 0.60 \\
\midrule
\multirow{4}{*}{Qwen3-32B + SFT} & Level 1 &84.83  &38.28  &99.97  &86.82  &100  &94.44  & 0.56 \\
 & Level 2 &60.5 &8  &96.54  &40.48  &100 &75  & 1.50 \\
 & Level 3 &17 &0.17  &63.75  &1.67  &100 &16.67  & 0.00 \\
 \rowcolor{bggreen}
 & Mean &\textbf{71.89}  &27.31  &\textbf{95.18}  &67.06 &\textbf{100} & 81.48 &  0.75 \\
\midrule
\multirow{4}{*}{Qwen3-32B + SFT + RL} & Level 1 &82.17 &43.28 &99.92 &93.89 &100 &100 & 0.61 \\
 & Level 2 &62.75 &14.25 &95.49 &64.28 &100 &91.67 & 1.86 \\
 & Level 3 &14.67 &0.17  &61.12  &1.67  &100 &16.67  & 0.00 \\
 \rowcolor{bggreen}
 & Mean &70.35 &\textbf{32.04} &94.62 &\textbf{77.06} &\textbf{100} &\textbf{88.89} & 0.87 \\

\bottomrule
\end{tabular}}
\end{table}

\begin{figure*}
    \centering
    \includegraphics[width=0.98\linewidth]{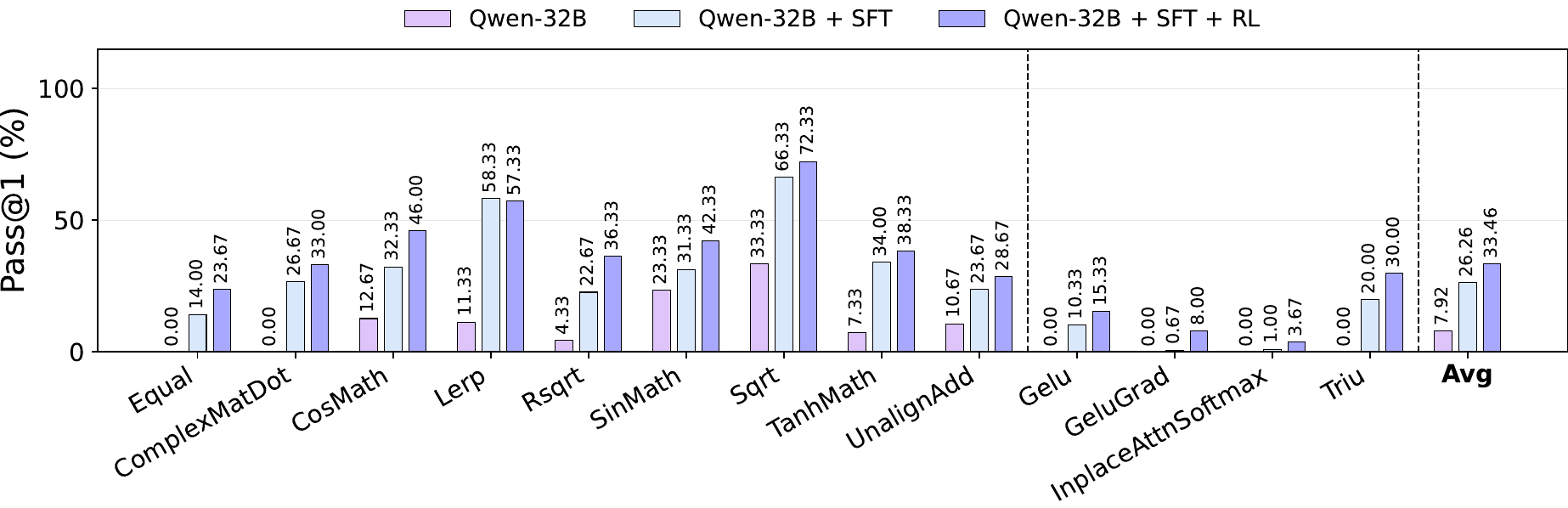}
    \caption{Pass@1 results for representative kernels across training stages.}
    \label{fig:Specific_kernel_acc}
\end{figure*}

\subsubsection{Compilation and Execution}

Table~\ref{tab:pass_k_results_speed} summarizes kernel generation performance on NPUKernelBench across different sampling budgets ($k$) and difficulty levels.
A clear performance gap emerges as kernel complexity increases.
For the base Qwen3-32B model, Level~1 kernels achieve moderate compilation rates (CR) and execution success rates (ER), whereas Level~2 kernels remain largely unsolved.
For Level~3 kernels, although higher $k$ improves CR to some extent, ER remains consistently low, indicating that successful compilation alone does not guarantee correct execution for complex kernels.

Figure~\ref{fig:Specific_kernel_acc} presents the Pass@1 accuracy of Qwen-32B on a set of representative kernels across three training stages: the base model, supervised fine-tuning (SFT), and SFT followed by reinforcement learning (RL).
The base model exhibits limited kernel synthesis capability, achieving an average Pass@1 of only 7.92\%, which reflects the difficulty of directly transferring general-purpose code generation ability to low-level NPU kernel programming.

Supervised fine-tuning leads to a substantial improvement, increasing the average Pass@1 to 26.26\%.
This gain indicates that SFT effectively aligns the model with the structural and syntactic requirements of kernel generation, enabling it to produce compilable and partially correct implementations.
In particular, SFT equips the model with essential building blocks such as API usage patterns, memory object initialization, and canonical computation templates, which are prerequisites for valid kernel synthesis.

Building upon this foundation, reinforcement learning further improves the average Pass@1 accuracy to 33.46\%.
Unlike SFT, which primarily addresses coarse-grained correctness, RL provides execution-based preference signals that encourage the model to distinguish between multiple valid but semantically different implementations.
The resulting improvements are especially pronounced for complex kernels, where subtle errors in memory staging, accumulation order, or synchronization often determine execution success.

Overall, the progressive performance gains across training stages highlight the complementary roles of SFT and RL.
SFT establishes baseline kernel generation competence by enforcing hard compilation and correctness constraints, while RL refines fine-grained execution logic beyond syntactic validity.
This staged alignment strategy proves crucial for improving first-attempt accuracy on challenging NPU kernels, where even minor implementation errors can lead to execution failure.

\subsubsection{Performance Speedup}

Beyond functional correctness, Table~\ref{tab:pass_k_results_speed} further evaluates the runtime efficiency of the generated kernels on real Ascend NPU hardware. 
The base model exhibits limited optimization capability, achieving only a $0.60\times$ speedup on Level~1 tasks and failing entirely on Level~2 and Level~3 kernels. 
This behavior indicates that general-purpose code generation lacks the ability to reason about hardware-specific parallelism and memory hierarchies.

In contrast, the proposed training pipeline yields substantial and systematic performance improvements. 
After supervised fine-tuning, the model not only recovers executable kernels but also learns performance-oriented implementation strategies. 
In particular, SFT achieves a $1.50\times$ speedup on Level~2 tasks, surpassing expert-written baselines. 
This result suggests that exposure to kernel-centric reasoning patterns enables the model to internalize optimized tiling, memory reuse, and operator fusion strategies that are difficult to infer from generic code corpora alone.

Reinforcement learning further refines these implementations by incorporating execution-based preference signals. 
On Level~2 kernels, RL improves the average speedup to $1.86\times$, while maintaining stable performance on Level~1 tasks ($0.61\times$). 
These gains primarily stem from fine-grained adjustments to memory access ordering, synchronization placement, and accumulation structure, rather than large-scale structural changes introduced during SFT.

Performance improvements are most pronounced on Level~2 tasks, where structured computation and localized data dependencies provide sufficient flexibility for optimization while remaining amenable to model-driven reasoning. 
In contrast, Level~1 kernels offer limited optimization space, and Level~3 kernels remain challenging due to global dependencies and complex control flow. 
Overall, these results indicate that the proposed SFT+RL pipeline enables the model to generate not only syntactically valid kernel code, but also hardware-efficient parallelization patterns, achieving expert-level and even expert-surpassing performance on a broad class of practical NPU operators.

\subsection{Ablation Analysis of Supervised Fine-tuning}

\subsubsection{Model Scale Sensitivity} 
\begin{figure}
    \centering
    \includegraphics[width=1\linewidth]{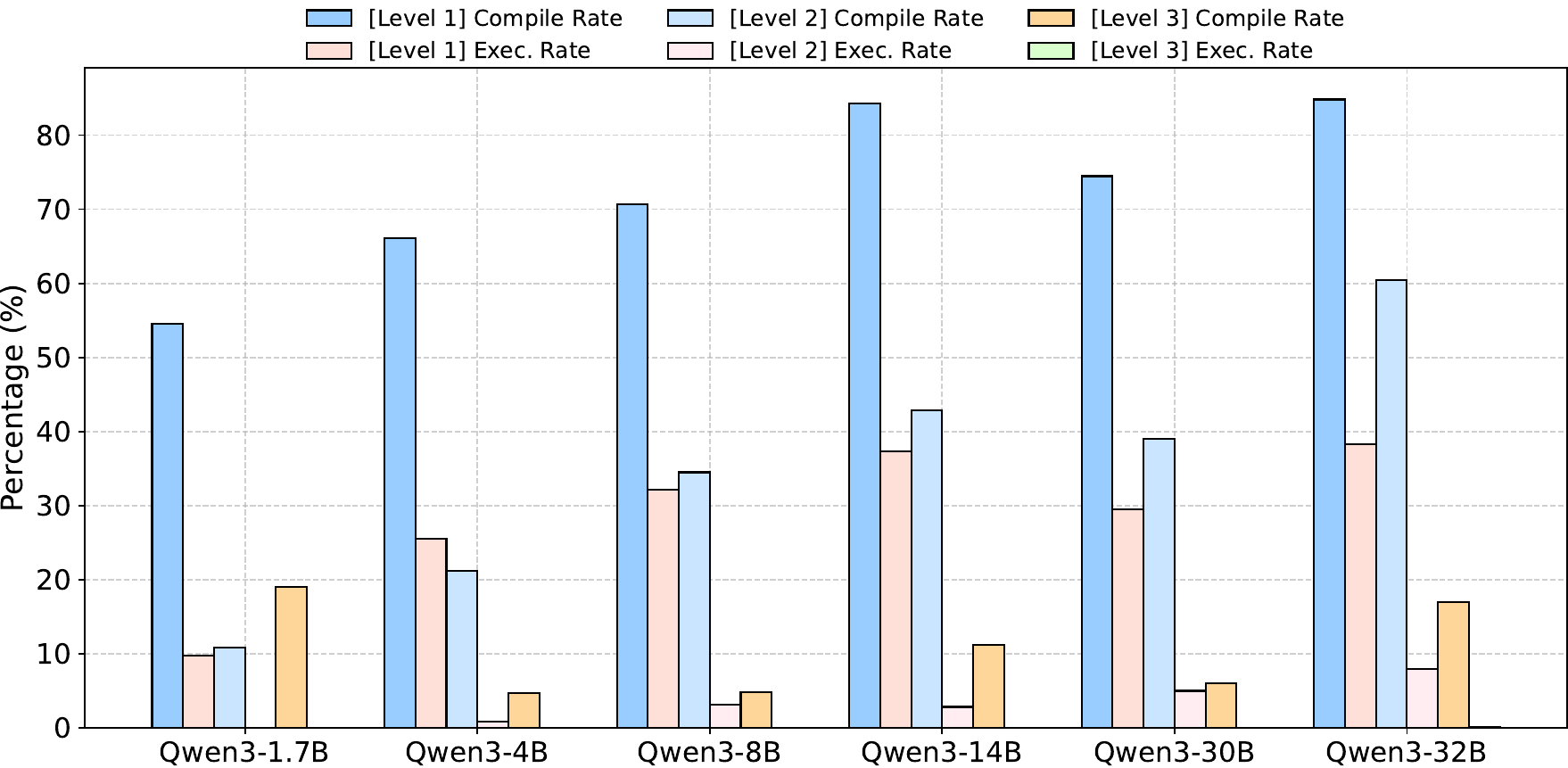}
    \caption{Scaling behavior of KernelGen-LM on NPU kernel generation tasks with increasing difficulty.}
    \label{fig:model_scale}
\end{figure}

As shown in Figure~\ref{fig:model_scale}, increasing model scale leads to clear performance gains on Level~1 and Level~2 tasks, indicating that larger models better capture standard kernel patterns. In contrast, Level~3 tasks remain highly challenging, with most models failing to achieve non-trivial execution rates. However, 32B is the only scale that demonstrates measurable compilation success on Level~3, suggesting that sufficient parameter scale is a prerequisite for reasoning over the complex control flow and memory dependencies of complicated NPU kernels.

\subsubsection{Comparison of Fine-tuning Strategies}  
\begin{table}[t]
    \centering
    \caption{Full-tuning \textit{V.S.} LoRA-tuning in QWen3-8B model.}
    \resizebox{0.75\linewidth}{!}{
    \begin{tabular}{l cc c c}
            \toprule
            \textbf{Model} & \textbf{Level} & \textbf{Compile Rate (\%)} & \textbf{Exec. Rate (\%)} & \textbf{Speedup ($\times$)} \\
            \midrule
            \multirow{4}{*}{LoRA-tuning}  & Level 1   & 54.94 & 20.10  & 0.48 \\  
            & Level 2    & 14.58 & 0.67   & 0.52 \\  
            & Level 3    & 3.83  & 0.00   & - \\ 
            \rowcolor{bggreen}
            & Mean       & 40.29 & 13.55  & 0.48 \\ \hline
             \multirow{4}{*}{Full-tuning}      & Level 1    & 70.67 & 32.17 & 0.58 \\
            & Level 2    & 34.50 & 3.08  & 2.77 \\ 
            & Level 3    & 4.83  & 0.00  & - \\ 
            \rowcolor{bggreen}
            & Mean       & \textbf{55.32} & \textbf{22.13} & \textbf{0.95} \\ 
            \bottomrule
        \end{tabular}
    \label{tab:full_vs_lora}}
\end{table}

Table~\ref{tab:full_vs_lora} compares a comparative analysis between full fine-tuning and LoRA-based fine-tuning on Qwen3-8B.
Full fine-tuning consistently outperforms LoRA across all kernel complexity levels, with the mean compilation rate surging from 40.29\% to 55.32\% and the execution rate improving from 13.55\% to 22.13\%.
The performance gap widens significantly as kernel complexity increases.
We attribute this divergence to the nature of NPU kernel generation, which is a knowledge-intensive task involving strict hardware intrinsics and memory constraints, rather than a simple style transfer.
While LoRA is effective for format alignment, its low-rank decomposition limits the capacity to absorb such high-dimensional, domain-specific knowledge.
In contrast, full fine-tuning updates the entire parameter space, allowing for the deep injection of complex semantics and logical dependencies required for valid code generation.
Consequently, in terms of runtime efficiency, full fine-tuning achieves an average speedup of $0.95\times$, substantially surpassing the $0.48\times$ of LoRA.
These results suggest that for precision-critical tasks like hardware-aware code generation, the comprehensive parameter update of full fine-tuning is indispensable.

\subsubsection{Training Data Composition Analysis} 
\begin{figure}[t]
    \centering
    \begin{subfigure}{0.552\textwidth}
        \centering
        \includegraphics[width=\linewidth]{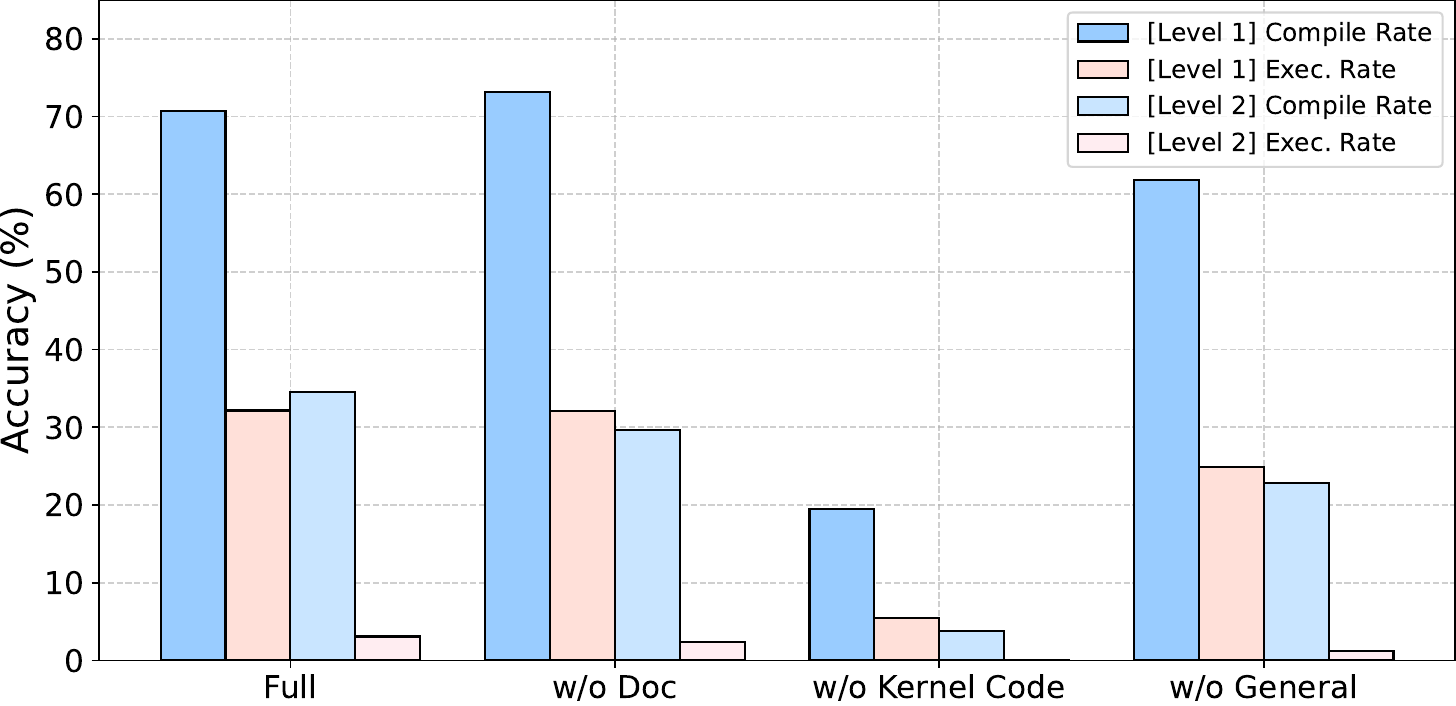}
        \caption{Pass@1 rates with different data compositions.}
        \label{fig:data_compos}
    \end{subfigure}
    \hfill
    \begin{subfigure}{0.438\textwidth}
        \centering
        \includegraphics[width=\linewidth]{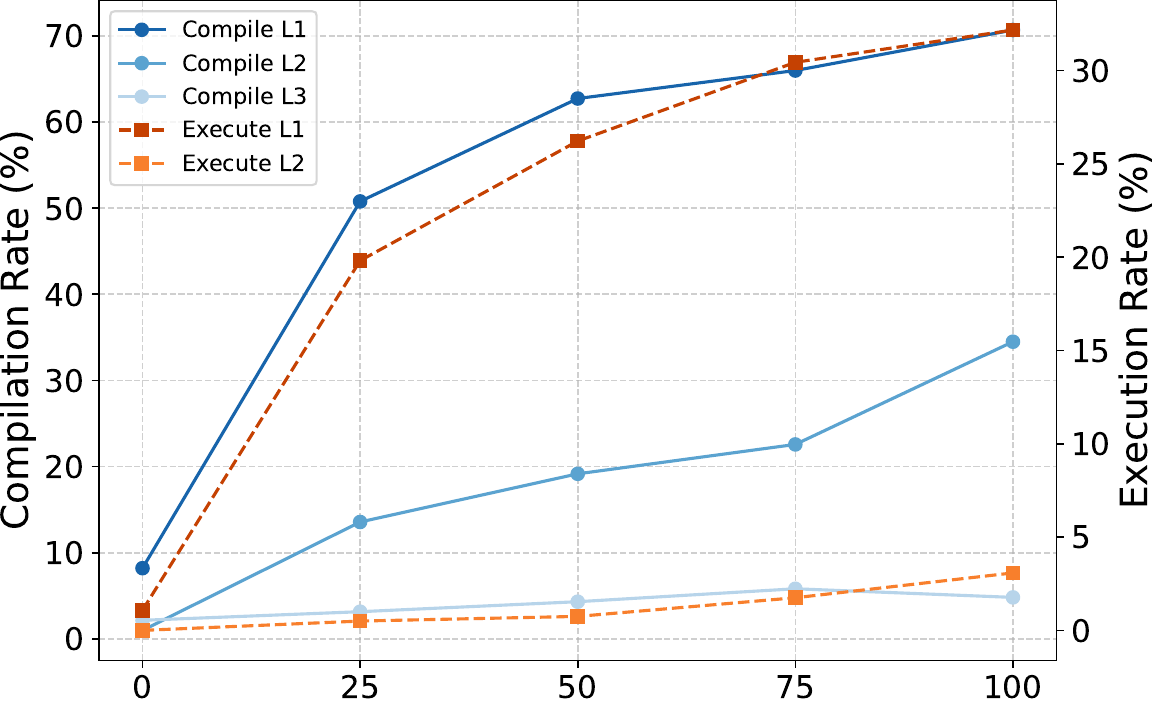}
        \caption{Pass@1 rates with varying data ratios in Full setting.}
        \label{fig:data_percent}
    \end{subfigure}
    \caption{Ablation results on training data composition. 
(a) Pass@1 compilation and execution rates for Level~1 and Level~2 kernels on Qwen3-8B under different data composition settings. Removing kernel code leads to the most severe performance degradation, particularly for Level~2 kernels.
(b) Pass@1 compilation and execution rates under the full data setting with different data ratios. Increasing the training data consistently improves both compilation and execution performance.}
    \label{fig:ablation}
\end{figure}

Fig.~\ref{fig:data_compos} shows that removing kernel code from the training data results in a dramatic drop in Pass@1 performance, underscoring its critical role.
Documentation and general semantic data provide auxiliary benefits, particularly for higher-complexity kernels.
Fig.~\ref{fig:data_percent} further demonstrates that increasing the training data size yields consistent improvements, highlighting the importance of data scale.

\subsection{Ablation Analysis of Reinforcement Learning}

\begin{table}[t]
\centering
\caption{DPO hyperparameter ablation in QWen3-8B.}
\resizebox{1\linewidth}{!}{
\begin{tabular}{l c c c c c c}
\toprule
\textbf{Negative Strategy} & \textbf{LR} & \textbf{LR Decay} & \textbf{Batch Size} & \textbf{Training Steps} & \textbf{Compile Rate (\%)} & \textbf{Exec. Rate (\%)} \\
\midrule
 (SFT) & - & - & - & - & \textbf{60.97} & \textbf{5.18}  \\
 \multirow{2}{*}{Compile-pass but execution-fail} 
  &  1e-6 &  constant &  64 &  150 & \textbf{44.21} & \textbf{8.31} \\
  & 1e-6 &  cosine &  64 &  150  & \textbf{54.03}  & \textbf{9.49}  \\
 Compile-fail &  1e-6 &  constant &  64 &  150 & \textbf{32.00}  & \textbf{6.10} \\
\bottomrule
\end{tabular}
}
\label{tab:DPO_hyper}
\end{table}

Table~\ref{tab:DPO_hyper} presents an ablation study on RL training with different negative strategies and optimization settings, where the negative strategies follow the preference data construction introduced in Sec.~\ref{reinforcement_learning}. Comparing the second row (compile-pass but execution-fail) with the fourth row (compile-fail) shows that using execution-failed yet compilation-passed samples as negatives leads to markedly better compilation and execution performance, indicating that such preference data provides more informative supervision than compile-failed negatives. Under the same negative strategy, a further comparison between the second and third rows demonstrates that cosine learning rate decay consistently outperforms a constant schedule, especially in terms of execution success. These results jointly motivate our choice of execution-failed but compilation-passed samples as negative preferences and cosine decay as the default optimization setting for RL training.

\subsection{Error Analysis}

\begin{wrapfigure}{r}{0.45\linewidth}
    \centering
    \includegraphics[width=\linewidth]{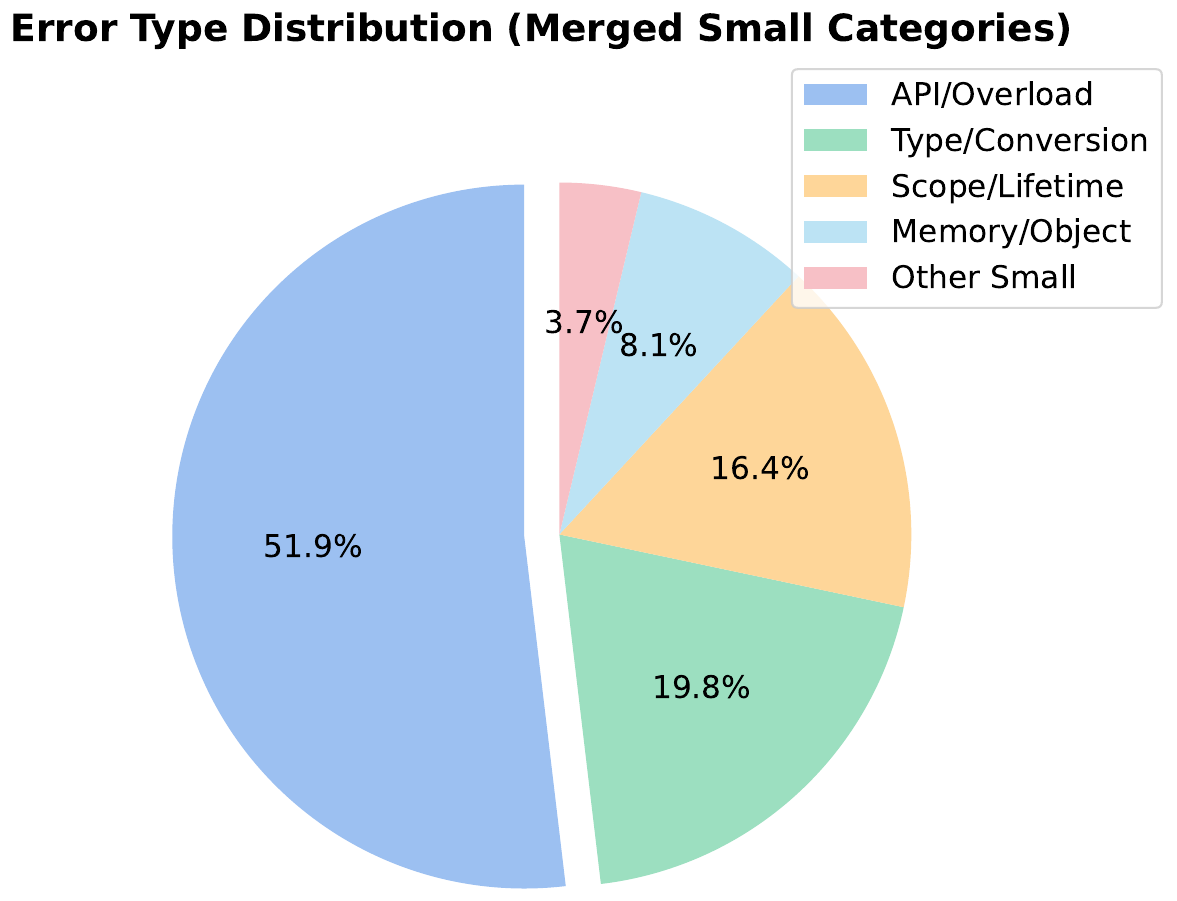}
    \caption{Distribution of kernel generation errors.}
    \label{fig:placeholder}
\end{wrapfigure}

To better understand the limitations of LLM-based kernel generation, we conduct a systematic error analysis over approximately 4,000 failed kernel generations.
Figure~\ref{fig:placeholder} summarizes the distribution of observed failure types.

As shown in the figure, API Signature and Overload Errors constitute the dominant failure mode, accounting for 51.9\% of all failed cases. These errors occur when generated function invocations fail to match any valid AscendC API signature or overload. The second most common category is Data Type and Conversion Errors (19.8\%), followed by Variable Scope and Lifetime Errors (16.4\%) and Memory and Object Usage Errors (8.1\%). Lower-frequency failures—including syntax and structural violations as well as macro and preprocessing errors—together account for only 3.7\% of all cases.

This distribution indicates that current models are generally capable of producing syntactically well-formed kernel code, as purely syntactic and structural errors are relatively rare.
However, the majority of failures arise from semantically constrained aspects of low-level programming.
In particular, errors are dominated by cases that require precise reasoning about API contracts, hardware-specific data types, and variable visibility across different kernel components.

To further investigate the underlying causes of these failures, we analyze the major error categories in detail. Below, we describe each error category along with its typical causes and manifestations.

\begin{itemize}
    \item \textbf{API Signature \& Overload Errors}: These errors arise when a generated function call does not match any valid API signature or overload defined in the programming interface. Common causes include incorrect argument ordering, mismatched parameter types, missing required parameters, or confusion between similarly named APIs with subtle semantic differences. This category reflects the model’s difficulty in aligning high-level task intent with rigid, hardware-specific API contracts.
    
    \item \textbf{Data Type \& Conversion Errors}: This category includes the use of invalid or non-existent data types (e.g., \texttt{bool\_t}), as well as illegal or unsafe type conversions between incompatible types (e.g., mixing \texttt{float} and \texttt{double} constants). Such errors often stem from implicit assumptions carried over from high-level programming languages, which do not hold in low-level kernel development where type rules are strict and hardware-dependent.

    \item \textbf{Variable Scope \& Lifetime Errors}: These errors occur when variables are used outside their valid scope or before being properly defined. Typical examples include references to undeclared symbols (e.g., \texttt{M\_PI}) or misuse of variables that are not visible within the current compilation unit or kernel context. This category highlights challenges in consistently reasoning about variable visibility and lifetime across host-side logic, kernel code, and auxiliary configuration structures.
    
    \item \textbf{Memory \& Object Usage Errors}: Errors in this category are related to incorrect usage of core objects such as \texttt{TPipe}, \texttt{LocalTensor}, and \texttt{GlobalTensor}. They include invalid member function invocations and incorrect memory address computations. While less frequent, such errors can lead to severe runtime failures or undefined behavior.

    \item \textbf{Syntax \& Structure Errors}: These errors stem from violations of C++ syntax or incomplete kernel class definitions. Examples include missing required methods (e.g., improperly defined or absent \texttt{Init} functions), mismatched braces, or malformed template instantiations.
    These errors are typically associated with incomplete code generation rather than deeper semantic misunderstandings.

    \item \textbf{Macro \& Preprocessing Errors}: This category includes errors caused by incorrect macro definitions or improper usage of tiling-related preprocessing directives. These issues often arise from incorrect assumptions about compile-time constants or conditional compilation logic embedded in kernel templates.

\end{itemize}

Representative examples for each error category are provided in Table~\ref{tab:error_examples} in the Appendix.
\section{Conclusion}
\label{conclusions}

In this work, we presented Ascend KernelGen, a unified framework designed to automate the generation of high-performance kernels for NPU architectures. We identified that the primary barrier for LLMs in this domain is not merely syntax familiarity, but the lack of structured reasoning capabilities regarding hardware constraints, memory hierarchy management, and asynchronous pipeline synchronization.

To overcome this, we constructed Ascend-CoT, a domain-specific dataset that explicitly models the reasoning process of expert developers, and developed NPUKernelBench, a rigorous evaluation suite that goes beyond static code analysis to verify compilation, numerical correctness, and runtime performance on actual hardware. Our extensive experiments show that while general-purpose models struggle with the specificity of AscendC, our domain-adaptive training strategy—combining reasoning-oriented supervised fine-tuning with execution-guided reinforcement learning—yields substantial improvements. We demonstrated that our model can successfully generate complex operators that were previously out of reach, achieving competitive performance metrics.

While this study focuses on kernel-level correctness and basic optimization, several promising directions remain. First, we plan to extend our framework to better support Level-3 kernels, which remain challenging for current models. Second, we aim to integrate performance-aware reward models into the reinforcement learning stage to prioritize not just functional correctness, but also latency minimization and resource utilization close to the hardware roofline. Finally, we intend to explore the generalizability of our methodology to other emerging accelerator platforms, fostering a more inclusive ecosystem for AI-driven hardware software co-design.
\clearpage
\begin{table}[t]
  \centering
  \begin{tcolorbox}[
    rounded corners,
    arc=3pt,
    enhanced,
    width=1\textwidth,
    colframe=refpurple!15, 
    coltitle=refpurple,
    colback=white,
    title=\textbf{Evaluation Criteria},
    fonttitle=\bfseries\small,
    boxrule=0.6pt,
    left=2pt,
    right=2pt,
    top=0pt,
    bottom=1pt
  ]
\begin{tcolorbox}[
      sharp corners,
      colframe=refgreen,
      colback=refgreen!5,
      colbacktitle=refgreen!15,
      coltitle=refgreen,
      boxrule=0.4pt,
      title={\textbf{Default evaluation criteria}},
      fonttitle=\bfseries\small,
      left=0pt,
      right=0pt,
      top=-4pt,
      bottom=-4pt
    ]
      \begin{center}
        \begin{lstlisting}[style=mystyle, frame=none,framerule=0pt, language=Python]
def check_precision(outputs, outputs_new, max_abs_error, max_rel_error):
    # outputs: list of tensors from reference model
    # outputs_new: list of tensors from LLM generated kernel
    outputs = [outputs] if not isinstance(outputs, list) else outputs
    outputs_new = [outputs_new] if not isinstance(outputs_new, list) else outputs_new
    
    all_abs_diff, all_rel_diff = [], []
    is_accurate = True
    
    # Process each output pair
    for out, out_new in zip(outputs, outputs_new):
      abs_diff = torch.abs(out - out_new)
      rel_diff = abs_diff / (torch.abs(out) + 1e-7) # Add epsilon to avoid division by zero
      all_abs_diff.append(abs_diff.view(-1))
      all_rel_diff.append(rel_diff.view(-1))
    
      # Check if any element exceeds both absolute and relative error thresholds
      if ((abs_diff > max_abs_error) & (rel_diff > max_rel_error)).any():
          is_accurate = False
    
    # Combine all differences
    all_abs_diff = torch.cat(all_abs_diff)
    all_rel_diff = torch.cat(all_rel_diff)
    
    return (1 if is_accurate else 0), all_abs_diff, all_rel_diff
\end{lstlisting}
\end{center}
\end{tcolorbox}

\begin{tcolorbox}[
      sharp corners,
      colframe=refgreen,
      colback=refgreen!5,
      colbacktitle=refgreen!15,
      coltitle=refgreen,
      boxrule=0.4pt,
      title={\textbf{Custom evaluation criteria}},
      fonttitle=\bfseries\small,
      left=0pt,
      right=0pt,
      top=-4pt,
      bottom=-5pt
    ]
      \begin{center}
        \begin{lstlisting}[style=mystyle, frame=none,framerule=0pt, language=Python,upquote=true]
def custom_check_precision(param, outputs, outputs_new):
    dtype_str = param.get(`dtype`, `float16`)
    dtype = getattr(torch, dtype_str)
    if dtype == torch.float32:
        return check_precision(outputs, outputs_new, max_abs_error=0.00001, max_rel_error=0.00001)
    else:
        return check_precision(outputs, outputs_new, max_abs_error=0.001, max_rel_error=0.001)
\end{lstlisting}
\end{center}
\end{tcolorbox}
\end{tcolorbox}
\vspace{-10pt}
\caption{Correctness evaluation criteria for LLM-generated kernels, covering both default and custom methods.}
\label{tab:kernel_correctness}
\end{table}

\begin{table}[H]

  \centering
  \begin{tcolorbox}[
    rounded corners,
    arc=3pt,
    enhanced,
    width=1\textwidth,
    colframe=codegreen, 
    colback=white,
    title=\textbf{Prompts for Generating Kernel Code in NPUKernelBench},
    fonttitle=\bfseries\small,
    boxrule=0.6pt,
    left=2pt,
    right=2pt,
    top=0pt,
    bottom=1pt
  ]
    \begin{tcolorbox}[
      sharp corners,
      colframe=reforange,
      colback=reforange!5,
      colbacktitle=reforange!15,
      coltitle=reforange,
      boxrule=0.4pt,
      title={\textbf{Instructions}},
      fonttitle=\bfseries\small,
      left=0.5pt,
      right=0.5pt,
      top=0.5pt,
      bottom=0.5pt
    ]
{\small Please see Table~\ref{tab:prompt_ins}.}
\end{tcolorbox}

\begin{tcolorbox}[
      sharp corners,
      colframe=taskblue,
      colback=taskblue!5,
      colbacktitle=taskblue!15,
      coltitle=taskblue,
      boxrule=0.4pt,
      title={\textbf{API\_Desc.md}},
      fonttitle=\bfseries\small,
      left=0pt,
      right=0pt,
      top=0pt,
      bottom=0pt
    ]
\begin{lstlisting}[style=mdstyle]
# aclnnBasicMatmul 
## Function Description 
### Kernel Function 
This Ascend C kernel performs the multiplication of two 2D matrices, A and B. It is a core component in deep learning models, such as linear layers and attention mechanisms. It takes two 2D tensors that satisfy the matrix multiplication rule and outputs their product.
### Computational Formula 
Assume the input tensors are $A$ ($m \times k$) and $B$ ($k \times n$). The output tensor $C$ ($m \times n$) is computed as: $$C_{ij} = \sum_{p=1}^{k} A_{ip} B_{pj}$$, where $i$ ranges from $1$ to $m$, and $j$ ranges from $1$ to $n$. $C_{ij}$ denotes the element in the $i$-th row and $j$-th column of $C$.
### Computation Process and Type Conversion
To maintain high numerical precision during large-scale accumulation and effectively prevent data overflow, this kernel adopts a high-precision accumulation strategy during computation. The process is as follows: 
1. The kernel receives two input tensors  `a` and `b`, both of data type `float16`.
2. During the multiply-accumulate computation, the internal accumulator uses the `float32` data type. In other words, the product results of `float16` inputs are first converted to `float32` before accumulation.
3. After all accumulation operations are completed, a result tensor of type `float32` is obtained.
4. Finally, the `float32` result tensor is converted back to `float16` as the final output.
## Interface Definition
### Kernel Prototype Definition Interface
#### Input 
- a: Device-side aclTensor corresponding to A in the formula; supports float16, 2D, and ND format.
- b: Device-side aclTensor corresponding to B in the formula; supports float16, 2D, and ND format.
#### Output 
- c: Device-side aclTensor corresponding to C in the formula; supports float16, 2D, and ND format.
#### Attr 
- None
## Constraints and Limitations
* The input tensors `a` and `b` currently support only the `float16` data type.
* The input tensors `a` and `b` must both be two-dimensional matrices.
* The second dimension (number of columns) of `a` must be equal to the first dimension (number of rows) of `b`.
* The input tensors support only the ND data format.
\end{lstlisting}
 \end{tcolorbox}
 
\begin{tcolorbox}[
      sharp corners,
      colframe=refgreen,
      colback=refgreen!5,
      colbacktitle=refgreen!15,
      coltitle=refgreen,
      boxrule=0.4pt,
      title={\textbf{Host template}},
      fonttitle=\bfseries\small,
      left=0pt,
      right=0pt,
      top=-4pt,
      bottom=-4pt
    ]
      \begin{center}
        \begin{lstlisting}[style=mystyle, frame=none,framerule=0pt, language=Python]
#include "register/op_def_registry.h" 
#include "tiling/platform/platform_ascendc.h" 
namespace optiling { 
static ge::graphStatus TilingFunc(gert::TilingContext *context) 
{ 
    context->SetBlockDim(platform_ascendc::PlatformAscendCManager::GetInstance()->GetCoreNumAic()); 
    return ge::GRAPH_SUCCESS; 
} 
} // namespace optiling 
namespace ops { 
class BasicMatmul : public OpDef { 
public: 
    explicit BasicMatmul(const char *name) : OpDef(name) 
{ 
        this->Input("a") 
            .ParamType(REQUIRED) 
            .DataType({ge::DT_FLOAT16}) 
            .Format({ge::FORMAT_ND}); 
        this->Input("b") 
            .ParamType(REQUIRED) 
            .DataType({ge::DT_FLOAT16}) 
            .Format({ge::FORMAT_ND}); 
        this->Output("c") 
            .ParamType(REQUIRED) 
            .DataType({ge::DT_FLOAT16}) 
            .Format({ge::FORMAT_ND}); 
        this->AICore() 
            .SetTiling(optiling::TilingFunc) 
            .AddConfig("ascend910_93") 
            .AddConfig("ascend910b"); 
    } 
}; 
OP_ADD(BasicMatmul); 
} // namespace ops
\end{lstlisting}
\end{center}
\end{tcolorbox}

\begin{tcolorbox}[
      sharp corners,
      colframe=refgreen,
      colback=refgreen!5,
      colbacktitle=refgreen!15,
      coltitle=refgreen,
      boxrule=0.4pt,
      title={\textbf{Kernel template}},
      fonttitle=\bfseries\small,
      left=0pt,
      right=0pt,
      top=-4pt,
      bottom=-5pt
    ]
      \begin{center}
        \begin{lstlisting}[style=mystyle, frame=none,framerule=0pt, language=Python]
#include <kernel_operator.h> 
using namespace AscendC; 
extern "C" __global__ __aicore__ void basic_matmul(GM_ADDR a, GM_ADDR b, GM_ADDR c, GM_ADDR workspace, GM_ADDR tiling) 
{ 
}
\end{lstlisting}
\end{center}
\end{tcolorbox}
\end{tcolorbox}
\vspace{-10pt}
\caption{An example of the structured prompt used for Ascend kernel generation, consisting of three components: API description, host/kernel-side template.}
\label{tab:kernel_gen}
\end{table}

\begin{table*}
\centering
    \begin{tcolorbox}[
    breakable,
      sharp corners,
      colframe=refpurple,
      colback=refpurple!5,
      colbacktitle=refpurple!15,
      coltitle=refpurple,
      boxrule=0.4pt,
      title={Kernel-level CoT}, 
      fonttitle=\bfseries\small,
      left=0.5pt,
      right=0.5pt,
      top=0.5pt,
      bottom=0.5pt
    ]

\begin{lstlisting}[language=json]
{
  "qus": "Please implement a utility function for computing the temporary buffer size required by the SwiGLU operator. The function should be able to determine the maximum and minimum buffer sizes based on the input shape and data type, as well as retrieve the buffer’s factor size. Some dependency code has already been provided; please complete the implementation based on the following summarized dependencies: [==========From swiglu\_tiling.h ==================...code...================== From /api\_check/host\_apicheck.h =======...code...==================== From impl/host\_log.h ======================...code...]",
  "cot": "To solve this problem, I need to implement a utility function that computes the temporary buffer size required by the SwiGLU operator. The function must determine the maximum and minimum buffer sizes based on the input shape and data type, as well as return the buffer-size factor. According to the provided dependency code, basic interfaces and parameter-validation utilities already exist, so my task is to implement the core logic. Since SwiGLU requires intermediate results, its temporary-space needs vary across data types, and the Ascend processor also imposes alignment constraints, motivating the use of a minimum repeat-unit size. My design sets different multipliers for half-precision and single-precision data, defines a 256-byte minimum repeat unit, and validates both shape and datatype. The maximum buffer size is computed as the multiplier times the larger of the input-tensor byte size and the repeat-unit size, while the minimum buffer size uses only the repeat-unit size. The factor-size interface simply returns the chosen multiplier, enabling callers to compute the exact temporary-space requirement dynamically.",
  "ans": "/* !\n * \\file swiglu_tiling.cpp\n * \\brief\n */\n# include \"lib/activation/swiglu_tiling.h\"\n\n#include <cstdint>\n#include <set>\n\n#include \"graph/tensor.h\"\n#include \"impl/host_log.h\"\n#include \"impl/api_check/host_apicheck.h\"\n\nnamespace AscendC {\nnamespace {\nconstexpr uint32_t SWIGLU_HALF_TMP_SIZE = 6;\nconstexpr uint32_t SWIGLU_FLOAT_TMP_SIZE = 0;\nconstexpr uint32_t SWIGLU_ONE_REPEAT_BYTE_SIZE = 256;\nstatic constexpr uint32_t SWI_GELU_HALF_SIEZ = 2;\nstatic constexpr uint32_t SWI_GELU_FLOAT_SIEZ = 4;\nstatic const std::set<uint32_t> SUPPORT_TYPESIZE = { SWI_GELU_HALF_SIEZ, SWI_GELU_FLOAT_SIEZ };\nstatic constexpr const char SWI_GELU_GET_MAX_MIN[] = \"GetSwiGLUMaxMinTmpSize\";\nstatic constexpr const char SWI_GELU_GET_TMP_BUFFER[] = \"GetSwiGLUTmpBufferFactorSize\";\n\ninline uint32_t GetSwiGLUMaxTmpSize(const uint32_t inputSize, const uint32_t typeSize)\n{\n    const uint32_t calcPro = (typeSize == sizeof(float)) ? SWIGLU_FLOAT_TMP_SIZE : SWIGLU_HALF_TMP_SIZE;\n    return calcPro * std::max(inputSize * typeSize, SWIGLU_ONE_REPEAT_BYTE_SIZE);\n}\n\ninline uint32_t GetSwiGLUMinTmpSize(const uint32_t typeSize)\n{\n    const uint32_t calcPro = (typeSize == sizeof(float)) ? SWIGLU_FLOAT_TMP_SIZE : SWIGLU_HALF_TMP_SIZE;\n    return calcPro * SWIGLU_ONE_REPEAT_BYTE_SIZE;\n}\n} // namespace\n\nvoid GetSwiGLUMaxMinTmpSize(const ge::Shape& srcShape, const uint32_t typeSize, uint32_t& maxValue,\n    uint32_t& minValue, const bool isReuseSource)\n{\n    HighLevelApiCheck::SrcShapeSizeVerifyingParameters<SWI_GELU_GET_MAX_MIN>(srcShape.GetShapeSize(), typeSize);\n    HighLevelApiCheck::TypeSizeVerifyingParameters<SWI_GELU_GET_MAX_MIN>(typeSize, SUPPORT_TYPESIZE);\n    HighLevelApiCheck::IsReuseSourceVerifyingParameters<SWI_GELU_GET_MAX_MIN>(isReuseSource);\n    const uint32_t inputSize = srcShape.GetShapeSize();\n    ASCENDC_HOST_ASSERT(inputSize > 0, return, \"Input Shape size must be greater than 0.\");\n\n    minValue = GetSwiGLUMinTmpSize(typeSize);\n    maxValue = GetSwiGLUMaxTmpSize(inputSize, typeSize);\n}\n\nvoid GetSwiGLUTmpBufferFactorSize(const uint32_t typeSize, uint32_t &maxLiveNodeCount, uint32_t &extraBuffer)\n{\n    HighLevelApiCheck::TypeSizeVerifyingParameters<SWI_GELU_GET_TMP_BUFFER>(typeSize, SUPPORT_TYPESIZE);\n    extraBuffer = 0;\n    maxLiveNodeCount = (typeSize == sizeof(float)) ? SWIGLU_FLOAT_TMP_SIZE : SWIGLU_HALF_TMP_SIZE;\n}\n} // namespace AscendC"
}
\end{lstlisting}
\end{tcolorbox}
\vspace{-10pt}
\caption{A SwiGLU kernel CoT example for Ascend kernel SFT, consisting of three parts: question, CoT, and answer.}
\label{tab:singel_example}
\end{table*}
\begin{tcolorbox}[
    breakable,
    rounded corners,
    arc=3pt,
    enhanced,
    width=1\textwidth,
    colframe=codegreen, 
    colback=white,
    title=\textbf{Prompt for generating project-level CoT},
    fonttitle=\bfseries\small,
    boxrule=0.6pt,
    left=2pt,
    right=2pt,
    top=0pt,
    bottom=1pt
  ]
    \begin{tcolorbox}[
      sharp corners,
      colframe=reforange,
      colback=reforange!5,
      colbacktitle=reforange!15,
      coltitle=reforange,
      boxrule=0.4pt,
      title={\textbf{Instructions}},
      fonttitle=\bfseries\small,
      left=0.5pt,
      right=0.5pt,
      top=0.5pt,
      bottom=0.5pt
    ]
{\small 
You are an AscendC programming expert for the Huawei Ascend processor. Based on the given problem, code content, and hardware specification documentation, please generate a detailed chain-of-thought process and a summary of all member variable values within the structures, as well as the value of the tilingKey.
The chain of thought should demonstrate the complete reasoning path from the problem to the final code outcome, including key thinking steps, logical analysis, and derivation.

Requirements:

Start reasoning from the problem itself, with reasoning details aligned with the implementation logic;

Present a clear solution path;

Include necessary analytical steps and intermediate reasoning;

Guide the reasoning process step by step toward the final answer;

Demonstrate your expertise and reasoning process as an AscendC specialist;

......

For the chain of thought, the following requirements apply:

1. Begin from the problem and reason in the first-person “I”.

2. Present yourself as a developer actively solving the problem, not as an AI that already knows the answer.

3. Show the natural thought process from problem analysis to solution design and then to implementation. Code snippets may only reproduce content from the given material and must not be invented. Avoid statements such as “based on the example”.

4. Include key decision points, technical considerations, and derivations.

......

11. Except for code and proper nouns, do not use English inside the internal reasoning process (chain of thought itself).

12. Tiling parameters mainly include the following components:

    - {BlockDim}: the value set by the \texttt{GetBlockDim()} function
    
    - {TilingKey}: the value set by the \texttt{SetTilingKey()} function (optional)
    
    - {Tiling structure member variables}: typically defined between
      \texttt{BEGIN\_TILING\_DATA\_DEF()} and \texttt{END\_TILING\_DATA\_DEF()} in \texttt{*\_tiling.h}
    
......

The chain of thought should be output in the following format:

\texttt{<think>chain-of-thought</think}
.}
\end{tcolorbox}

\begin{tcolorbox}[
      sharp corners,
      colframe=taskblue,
      colback=taskblue!5,
      colbacktitle=taskblue!15,
      coltitle=taskblue,
      boxrule=0.4pt,
      title={\textbf{Test cases}},
      fonttitle=\bfseries\small,
      left=0pt,
      right=0pt,
      top=0pt,
      bottom=0pt
    ]
\begin{lstlisting}[style=mdstyle]
## Function Description
### Operator Function
Returns a new tensor with the same shape as the input, where each element is the absolute value of the corresponding input element.
### Formula
$$
x = [x_0, x_1, \ldots, x_{n-1}] \\
y = [y_0, y_1, \ldots, y_{n-1}]
$$
$$
y_i = |x_i|,\quad i = 0, 1, \ldots, n-1
$$
## API Definition
### Python API
This operator is implemented in C++ and exposed to Python via PyBind11 as `kernel_gen_ops.abs_math()`:
```python
def abs_math(tensor):
    """
    Custom AbsMath operator.

    Args:
        tensor (Tensor): Device-side aclTensor (input x).
            Supported dtypes: bfloat16, float16, float32, int32, int64, DT_COMPLEX64.
            Supported format: ND.

    Returns:
        Tensor: Device-side aclTensor (output y).
            Same dtype/format as input. Output shape matches x.
    """
```
## Usage Example
```python
import torch
import kernel_gen_ops

tensor = torch.tensor([-1.0, 2.0, -3.0], dtype=torch.float)
result = kernel_gen_ops.abs_math(tensor)
```
## Constraints
- ND format only.
\end{lstlisting}
 \end{tcolorbox}
 
\begin{tcolorbox}[
      sharp corners,
      colframe=refgreen,
      colback=refgreen!5,
      colbacktitle=refgreen!15,
      coltitle=refgreen,
      boxrule=0.4pt,
      title={\textbf{NPUs hardware configuration}},
      fonttitle=\bfseries\small,
      left=0pt,
      right=0pt,
      top=-4pt,
      bottom=-4pt
    ]
      \begin{center}
        \begin{lstlisting}[style=mdstyle]
The introductory document for AscendC development includes the following sections:
1. Overview of the Decoupled Architecture and Hardware Structure
1.1 Features of the Decoupled Architecture
The decoupled architecture of Ascend AI processors (e.g., Atlas A2/A3 series) splits the AI Core into two independent units: AI Cube (AIC) and AI Vector (AIV).
AIC mainly handles matrix operations (Cube), while AIV is responsible for vector and scalar operations (Vector/Scalar).
The two types of cores have independent scalar units, instruction streams, and local storage. Data is exchanged through Global Memory (GM).
......
The hardware model and specifications we are using are: [Platform Info configuration begin]
#*************************************************************************************#
[
[version]
SoC_version=Ascend910_9392
Short_SoC_version=Ascend910_93   [PS: Ascend910_93 is the NPU's name.]
AIC_version=AIC-C-220
CCEC_AIC_version=dav-c220-cube
CCEC_AIV_version=dav-c220-vec
CCEC_CUBE_version=dav-c220-cube
CCEC_VECTOR_version=dav-c220-vec
......
]

\end{lstlisting}
\end{center}
\end{tcolorbox}

\begin{tcolorbox}[
      sharp corners,
      colframe=refgreen,
      colback=refgreen!5,
      colbacktitle=refgreen!15,
      coltitle=refgreen,
      boxrule=0.4pt,
      title={\textbf{Kernel source code}},
      fonttitle=\bfseries\small,
      left=0pt,
      right=0pt,
      top=-4pt,
      bottom=-5pt
    ]
      \begin{center}
        \begin{lstlisting}[style=mdstyle]
Please, using the input shapes and other relevant information and taking the actual code implementation into account, outline the detailed design logic for operator tiling and kernel pipelining, and compute the operator’s tiling parameter values during the intermediate steps. Below is the code, consisting of multiple files separated by
============================================================================================.
You may refer to these files to support your reasoning, but they are only supplementary. Your answer should be based on your own thinking and judgment as an AscendC expert, not a verbatim restatement of the code:
============================================================================================
```cpp
// op_host/abs_tiling.h
/**
 * @file abs_tiling.h
 */
#ifndef ABS_TILING_H
#define ABS_TILING_H
#include "register/tilingdata_base.h"
namespace optiling {
    ......
}
#endif // ADD_CUSTOM_TILING_H
```
==========================================================================================
```cpp
// op_host/abs_math.cpp
/**
 * @file abs.cpp
 */
#include "abs_tiling.h"
#include "register/op_def_registry.h"
#include "tiling/platform/platform_ascendc.h"
namespace optiling {      
    ......
}
namespace ops {
class AbsMath : public OpDef {
    ......
};OP_ADD(AbsMath);
}
```
==========================================================================================
```cpp
// op_kernel/abs_common.h
/**
 * @file abs_common.h
 */
#ifndef ABS_COMMON_H
#define ABS_COMMON_H
#include "kernel_operator.h"
constexpr int32_t BUFFER_NUM = 2; // tensor num for each queue
#endif // ABS_COMMON_H
    ......
```
==========================================================================================
\end{lstlisting}
\end{center}
\end{tcolorbox}
\end{tcolorbox}
\vspace{-10pt}
\captionof{table}{An example of an AbsMath kernel prompt for Ascend kernel generation, comprising four components: instruction text, test cases, NPU hardware configuration, and kernel source code.}
\label{tab:prompt_project_file}
\clearpage
\begin{table}[t]
  \centering
    \begin{tcolorbox}[
    breakable,
      sharp corners,
      colframe=refpurple,
      colback=refpurple!5,
      colbacktitle=refpurple!15,
      coltitle=refpurple,
      boxrule=0.4pt,
      title={Project-level CoT}, 
      fonttitle=\bfseries\small,
      left=0.5pt,
      right=0.5pt,
      top=0.5pt,
      bottom=0.5pt
    ]

\begin{lstlisting}[language=json]
{
 "input": "Prompt in Table 9.",
 "output": "<think>\n My task is to implement an `AbsMath` operator for the Ascend processor. The operator's function is simple: take the absolute value of each element of the input tensor, i.e., `y = |x|`.\n\n First, let's analyze its characteristics. It is a typical element-wise operation with simple logic, no complex dependencies, and no matrix multiplication. Therefore, it is vector-intensive, meaning on the Ascend processor, I should mainly use the AI Vector (AIV) units rather than AI Cube (AIC) units.\n\nThe hardware model is `Ascend910_93` with `vector_core_cnt=48` Vector cores. Each core has a Unified Buffer (UB) of 196,352 bytes. UB size limits the amount of data that can be processed at once. The input is `shape=[1]`, `dtype=float16`. Though very small, the implementation (especially tiling) must handle arbitrary sizes. Based on this, the implementation is divided into Host-side Tiling logic and Device-side Kernel.\n\n**I. Host-side Tiling Strategy Design**\n\nThe core goal of Tiling is to decompose the entire computation task into small tiles so that it can be processed in parallel on multiple cores and ensure that each data block can be loaded into UB. \n\n1. **Data Type and Computation Selection**:\n The `AbsMath` operator needs to support multiple data types, each with different methods for calculating its absolute value. To differentiate these cases on the kernel side, I need to design a `tilingKey`. The host side sets this `tilingKey` based on the input data type, and the kernel side uses it to call the corresponding processing logic. * For `float16`, `bfloat16`, and `float32` types, the `Abs` instruction can be used directly, and I have assigned a `tilingKey` for this. * For `int32`, the absolute value needs to be efficiently obtained using the bitwise operation `(x ^ (x >> 31)) - (x >> 31)`, which corresponds to a separate `tilingKey`. * For `int64`, the bitwise operation `(x ^ (x >> 63)) - (x >> 63)` is also used, and a `tilingKey` is also assigned ...
}
\end{lstlisting}
\end{tcolorbox}
\captionof{table}{A project-level CoT example for Ascend kernel SFT, comprising a question and its CoT.}
\label{tab:project_example}
\end{table}

\begin{table}[t]
\vspace{-10pt}
\centering
\begin{tcolorbox}[
      sharp corners,
      colframe=reforange,
      colback=reforange!5,
      colbacktitle=reforange!15,
      coltitle=reforange,
      boxrule=0.4pt,
      title={\textbf{Prompts to optimize CoT}},
      fonttitle=\bfseries\small,
      left=0.5pt,
      right=0.5pt,
      top=0.5pt,
      bottom=0.5pt
    ]
{\small
"System Prompt":
You are a top-tier AI chain-of-thought optimization expert. Your primary task is to review and restructure the content inside the \texttt{<think>} tags in Markdown documents. You must precisely preserve the core reasoning path from problem analysis to code implementation, while removing all redundant summarizing text, parameter recap lists, and any non-essential reasoning. Ultimately, you must output only the optimized, complete Markdown document.

"User Prompt":

\textbf{Optimization Goals}

Reconstruct the content inside the \texttt{<think>} tags to retain the core reasoning path from problem analysis to code implementation, while deleting all non-essential and redundant information.

\textbf{Content Editing Rules}

1. \textbf{Remove redundant summaries}: You must delete all retrospective or summarizing text about \textbf{Tiling}, such as:
    \begin{itemize}
    \setlength{\itemsep}{0pt}
    \setlength{\parskip}{0pt}
    \setlength{\topsep}{0pt}
        \item ``Tiling parameter summary''
        \item ``Organize all tiling parameters''
        \item  Any similar parameter recap or retrospective content.
    \end{itemize}

2. \textbf{Remove self-negation}: You must delete all expressions of doubt, alternative-scheme discussion, or self-correction. The chain of thought should present a clear and confident design path.

3. \textbf{Enhance pipeline design using the provided Kernel code}.

4. \textbf{Do not modify or add any content outside the above three items.}

\textbf{Structural Adjustment Rules}

1. \textbf{Keep the single best block}: If multiple \texttt{<think>...</think>} blocks exist, you must keep only the most complete and clear one, and remove all others.

\textbf{Input and Output}
\begin{itemize}
\setlength{\itemsep}{0pt}
    \setlength{\parskip}{0pt}
    \setlength{\topsep}{0pt}
    \item \textbf{Input-1 (Kernel Code)}: The Kernel code section below, used to assist your understanding for supplementing pipeline design.
    \item \textbf{Input-2 (Original Markdown)}: The CoT that must be processed.
    \item \textbf{Output}: Return the fully optimized Markdown document directly. Do not include any additional explanations or annotations.
\end{itemize}
}
\end{tcolorbox}
\captionof{table}{General instructions used in the prompt.}
\label{tab:optimization_prompt}
\end{table}
\clearpage
\begin{table}[t]
\centering
\caption{Representative examples of errors in generated kernels, grouped by error type.}
\label{tab:kernel_errors_full}
\begin{tabular}{p{5cm} p{2cm} p{8cm}}
\toprule
\textbf{Error Type} & \textbf{Example} & \textbf{Description} \\
\midrule
\multirow{5}{*}{API Signature and Overloading} 
& Equal & Muls: third argument LocalTensor<float> instead of scalar; Greater: function not declared; Mins: third argument minValLocal LocalTensor<float> instead of float; DataCopy: LocalTensor<float> vs GlobalTensor<half> type conflict; ReduceSum: missing required template parameter \texttt{pattern}. \\
& Arange & Add: 4 arguments including float16 scalar, but all overloads require LocalTensor and more parameters; Duplicate: 3 arguments (LocalTensor, int, int) provided, but all overloads require 6 arguments or have type conflicts (unsigned int vs int). \\
\midrule
\multirow{2}{*}{Data Type and Casting} 
& Equal & equal: float to bool conversion unsupported by CastIntrinsicsImpl; Duplicate: target yLocal type LocalTensor<bool> conflicts with template float; Sub: input tensors type inference conflict (bool vs float). \\
& Less & Duplicate used with uint8\_t/bool, unsupported types; int16\_t to bool conversion missing CastIntrinsicsImpl overload. \\
\midrule
\multirow{2}{*}{Memory and Object Misuse} 
& Icamax & TQue missing GetSize; undefined FLT\_MAX; undefined fabs; another TQue missing GetSize; DataCopy: parameter mismatch, expected (GlobalTensor, LocalTensor, int); Init: 4 arguments provided, 5 expected. \\
& Isamax & Dereferencing uint64\_t from LocalTensor::GetPhyAddr(); DataCopy: reinterpret\_cast int32\_t* to \_\_gm\_\_ int32\_t* invalid. \\
\midrule
\multirow{2}{*}{Variable Scope and Lifetime} 
& IsInf & Undeclared identifier 'IsInf'; Undeclared 'ToLocalTensor', likely meant 'LocalTensor', used in And function. \\
& Ccopy & Undeclared 'complex64' in GlobalTensor/LocalTensor templates and sizeof; Undeclared 'c10', causing template instantiation and sizeof errors. \\
\midrule
\multirow{2}{*}{Syntax and Structural} 
& Snrm2 & SyncAll() defined without parameters, called as template SyncAll<true>(); KernelSnrm2 missing member 'tailBlock'; tiling struct registration fails. \\
& FastGeluGrad & Variable name starting with digit ('1\_702\_x'), invalid in C++. \\
\midrule
\multirow{2}{*}{Macro and Preprocessing} 
& ClipByValue & TILING\_KEY\_IS macro misused; macro expansion may lack semicolon. \\
& Sasum & 'ALIGN\_SIZE' ambiguous (user-defined 8 vs AscendC 32); DataCopyPad parameter mismatch; TBuf missing SetFlag member function. \\
\bottomrule
\end{tabular}
\label{tab:error_examples}
\end{table}

\bibliography{anthology}

\end{document}